\documentclass[journal]{IEEEtran}
\ifCLASSINFOpdf
\else
\fi
\usepackage{cite}
\usepackage{algorithmic}
\usepackage{graphicx}
\usepackage{textcomp}
\usepackage{epstopdf}
\usepackage{booktabs}
\usepackage{amsmath}
\usepackage{hyperref}
\usepackage{tipa}
\usepackage{esint}
\usepackage{arydshln}
\usepackage{multirow}
\usepackage{cite}
\usepackage{amssymb}
\usepackage{array}
\usepackage{color}
\usepackage{colortbl}
\usepackage{float}

\usepackage[numbers,sort&compress]{natbib}
\usepackage[table,xcdraw]{xcolor}
\usepackage{caption}
\begin{document}
%
\title{ D-Net: A Dual-encoder Network for Image Splicing Forgery Detection and Localization}
%
%
%

\author{
	    Bo Liu,
        Ranglei Wu,
        Xiuli Bi
        Bin Xiao,
        Weisheng Li,~\IEEEmembership{Member,~IEEE,}
        Guoyin Wang,~\IEEEmembership{Senior~Member,~IEEE}
        and Xinbo Gao,~\IEEEmembership{Senior~Member,~IEEE}}

\maketitle

\begin{abstract}
Recently, many detection methods based on convolutional neural networks (CNNs) have been proposed for image splicing forgery detection. Most of these detection methods focus on the validation of local patches or local objects. We regard image splicing forgery detection as a binary classification task that distinguishes tampered and non-tampered regions by forensic fingerprints rather than semantic features. As the network goes deep, its representation ability becomes strong. However, the non-semantic forensic fingerprints can hardly be retained by normal CNNs in deep layers. We proposed a novel dual-encoder network (D-Net) for image splicing forgery detection to resolve these issues, which employs an unfixed encoder and a fixed encoder. The unfixed encoder autonomously learns the image fingerprints that differentiate between the tampered and non-tampered regions, whereas the fixed encoder intentionally provides prior structural information that assists the learning and detection of the forgeries. This dual-encoder is followed by a spatial pyramid global-feature extraction module that expands the global insight of D-Net for classifying the tampered and non-tampered regions more accurately. In an experimental comparison study of D-Net and state-of-the-art methods, D-Net, without pre-training or training on a large number of forgery images, outperformed the other methods in pixel-level forgery detection. Moreover, it is stably robust to different anti-forensic attacks.
\end{abstract}

\begin{IEEEkeywords}
image splicing forgery detection, tampered region localization, convolutional neural network, dual-encoder, image fingerprints, global perspective
\end{IEEEkeywords}

%
\IEEEpeerreviewmaketitle

\section{Introduction}
\label{1}

\begin{figure}[t]
	\centering
	\includegraphics[scale=0.75]{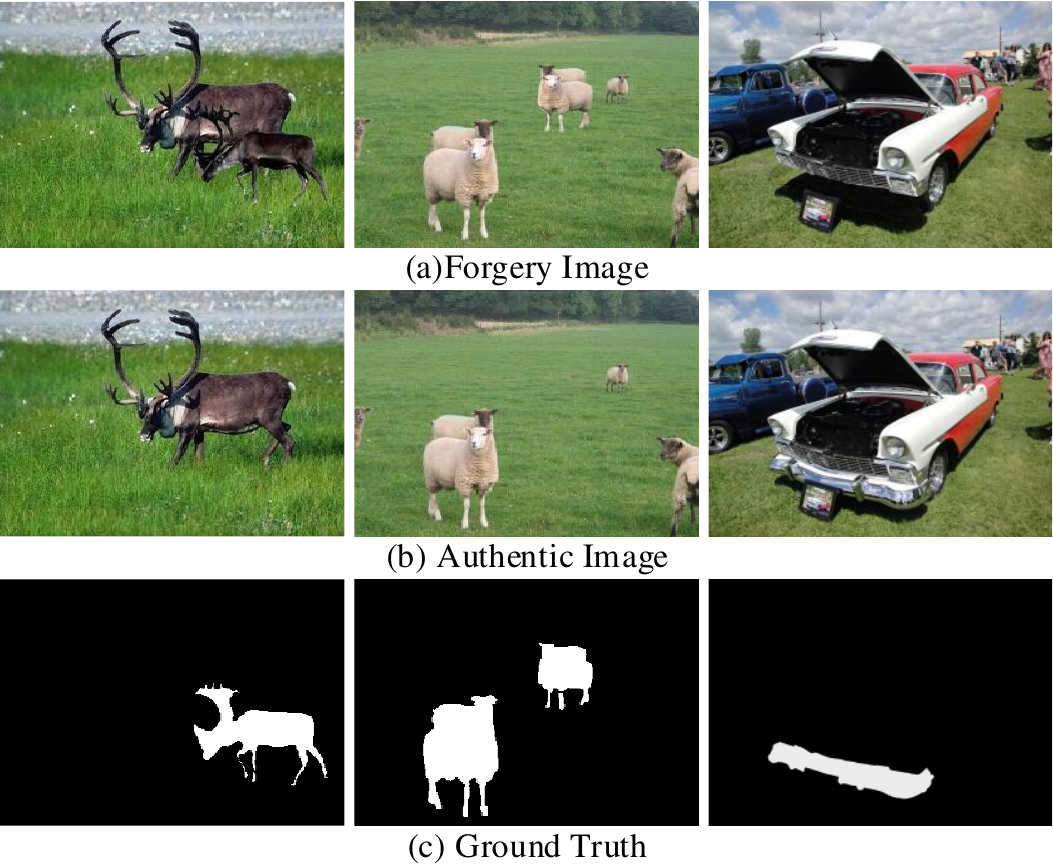}
	\caption{Examples of image local forgery. The first, second and third columns show an image formed by splicing forgery, an image formed by a copy-move forgery, and an image formed by a removal forgery, respectively. Rows \emph{a}, \emph{b} and \emph{c} display the forgery image, the authentic image and the ground truth, respectively.}
	\label{fig1}
\end{figure}
Discriminating between a tampered and non-tampered image is a challenging task in modern times because powerful image processing software and applications can easily manipulate images while leaving no visual clues of the manipulation. Therefore, confirming the authenticity of an image is an increasingly critical research topic. Depending on whether the semantic content of the image is changed or unchanged, image forgery is roughly divisible into two categories: global forgery \cite{kao2012effective,cao2014contrast,yerushalmy2011digital,chen2017blind} and local forgery. Global forgery is considered less harmful because it changes only the image's visual effect (by techniques such as image compression, image blurring, and image contrast enhancement) without altering the semantic content. On the contrary, a local forgery alters the semantic content of the image, as shown in Fig. \ref{fig1}. Because a local forgery can propagate false or misleading information through manipulated content, it is more harmful in different situations than global forgery. For local forgery, copy-move and removal forgery manipulate a single source image, but splicing forgery copies and pastes regions from one or more source images onto the target image ($1^{st}$ column of Fig. \ref{fig1}). The paper focuses on detecting the tampered regions in splicing-forged images.

All images possess intrinsic properties caused by imaging processing or post-processing. These intrinsic properties are unrelated to the image contents, and they can be exploited to differentiate an image from any other image. For this reason, they are called \emph{image fingerprints}. In splicing forgery images, the tampered and non-tampered regions come from different source images. Accordingly, a splicing forgery can be detected by finding different image fingerprints in different regions of an image. During the last decade, two main categories of image-fingerprint detection have been proposed: those based on traditional feature extraction and convolutional neural networks (CNNs). A traditional detection method extracts a particular image fingerprint, such as image-compression attributes \cite{lin2009fast,johnson2007exposing}, camera characteristics \cite{hsu2006detecting}, edge inconsistencies \cite{fang2010image,qu2009detect}, and photo-response nonuniformity noise \cite{chen2008determining,pan2011exposing,lyu2014exposing}. Because they target one fingerprint, such methods fail when the targeted fingerprint is missing or nonobvious. Moreover, post-processing operations such as image blurring, JPEG compression, and down-sampling will affect the particular image fingerprint, which degrades the detection results of traditional detection methods.

The use of CNNs in image-splicing forgery detection is inspired by the huge success of CNNs in computer vision. Table \ref{tab1} summarizes the notable CNN-based detection methods for image splicing forgery proposed since 2016. In the table, these methods are classified in two dimensions: (1)performing splicing detection on images or image patches; (2)learning image fingerprints intentionally or autonomously.
\begin{table}[ht]
	\setlength{\abovecaptionskip}{0.1 cm}
	\setlength{\belowcaptionskip}{6 pt}
	\centering
	\caption{Summary of CNN-Based Image Forgery Detection Methods that Realize Pixel-Level Localization and Perform Image-Level Classification (marked with $\diamond$).}
	\renewcommand\arraystretch{1.5}
	\setlength{\tabcolsep}{2pt}{
		\begin{tabular}{cc||m{3.6cm}<{\centering}|m{3.3cm}<{\centering}}
			\multicolumn{2}{c||}{}                    &\multicolumn{2}{c}{\small \textbf{Performing splicing detection}}        \\
			\multicolumn{2}{c||}{}                    &\multicolumn{1}{c|}{\small \textbf{based on image patches}}&\small \textbf{based on images}\\ \hline \hline
			&   &                                                                     &                                                \\
			&   &                                                                     &                                                \\
			&   &                                                                     &                                                \\
			&   &                                                                     &                                                \\
			& \multirow{-5}{*}{\rotatebox{90}{\small \textbf{autonomously}}} &\multirow{-5}{*}{\begin{tabular}[c]{@{}c@{}}{Zhang, \emph{et al}. \cite{zhang2016image}(2016)}\\ {Wei, \emph{et al}. \cite{wei2018c2r}(2018)}\\ {Cozzolino, \emph{et al}. \cite{cozzolino2019noiseprint}(2019)} \\ {Xiao, \emph{et al}. \cite{xiao2020image}(2020)} \end{tabular}} & \multirow{-5}{*}{\begin{tabular}[c]{@{}c@{}}$\diamond$Bayar, \emph{et al}. \cite{bayar2016deep}(2016)   \\ {Salloum, \emph{et al}. \cite{salloum2018image}(2018)} \\ $\diamond$Bayar, \emph{et al}. \cite{bayar2018constrained}(2018) \\ { Bi, \emph{et al}. \cite{bi2019rru}(2019)}\end{tabular}}      \\ \cline{2-4}
			& &                                                                     &                                                  \\
			& &                                                                     &                                                \\
			& &                                                                     &                                                \\
			& &                                                                     &                                                \\
			\multirow{-10}{*}{\rotatebox{90}{\small \textbf{Learning image fingerprints}}} & \multirow{-5}{*}{\rotatebox{90}{\small \textbf{intentionally}}} &\multirow{-5}{*}{\begin{tabular}[c]{@{}c@{}}$\diamond$Rao, \emph{et al}. \cite{rao2016deep}(2016) \\ {Bappy, \emph{et al}. \cite{bappy2017exploiting}(2017)} \\ {Shi, \emph{et al}. \cite{shi2018image}(2018)}\\ {Huh, \emph{et al}. \cite{huh2018fighting}(2018)} \\ {Liu, \emph{et al}. \cite{liu2020exposing}(2020)}\end{tabular}}     & \multirow{-5}{*}{\begin{tabular}[c]{@{}c@{}}$\diamond$Zhou, \emph{et al}. \cite{zhou2018learning}(2018)\\ {Bappy, \emph{et al}. \cite{bappy2019hybrid}(2019)}\\ {Wu, \emph{et al}. \cite{wu2019mantra}(2019)} \\{Zhang, \emph{et al}. \cite{zhang2020dense}(2020)}\\ $\diamond$Marra, \emph{et al}. \cite{marra2020full}(2020)  \end{tabular}}  \\
	\end{tabular}}
	\label{tab1}
\end{table}

Some researchers \cite{zhang2016image,wei2018c2r,cozzolino2019noiseprint,xiao2020image,rao2016deep,bappy2017exploiting,shi2018image,huh2018fighting,liu2020exposing} have regarded image splicing forgery detection as a binary classification on image patches. As these detection networks focus only on the local patches and ignore the relationship between the image patches, they easily return false decisions. For example, an image patch in the tampered regions will be predicted as a non-tampered patch because it shows a consistent image fingerprint within the patch. Moreover, as the final detection results are derived from the decisions of image patches, the detected regions are generally composed of square white blocks or the boundaries of the tampered regions. To resolve these problems, some researchers  \cite{bayar2016deep,bayar2018constrained,salloum2018image,bi2019rru,zhou2018learning,bappy2019hybrid,wu2019mantra,zhang2020dense,marra2020full} have employed end-to-end networks, which handle the image splicing forgery detection as an image segmentation or object detection task. As these methods consider the contextual relationships between local patches, they outperform patch-wise detection networks. We regard the image splicing forgery detection as a binary classification task distinguishing between tampered and non-tampered regions based on forensic features rather than semantic information. Especially, a single splicing forgery image may include both tampered semantic objects and tampered non-semantic regions, so the decisions of the detection network must cover these local objects and regions. However, the existing detection methods usually target either local patches or local objects, which is insufficient for high-performance splicing forgery detection. 

For autonomous learning image fingerprints, some detection methods \cite{zhang2016image,wei2018c2r,cozzolino2019noiseprint,xiao2020image,bayar2016deep,bayar2018constrained,salloum2018image,bi2019rru} have exploited the learning ability of CNNs or re-deployed existing CNNs originally proposed for other tasks. Whereas some researchers \cite{rao2016deep,bappy2017exploiting,shi2018image,huh2018fighting,liu2020exposing} have employed CNNs to learn only a particular image fingerprint, others \cite{zhou2018learning,bappy2019hybrid,wu2019mantra,zhang2020dense,marra2020full} have strengthened the autonomous learning of image fingerprints using a particular image fingerprint. Although these CNN-based detection methods have demonstrated high-performance learning of image fingerprints, their accuracy would be improved by considering specific image contents. For example, the structural information of the pixels, which may represent the boundaries of tampered regions, is an important trail but is almost ignored by CNNs. Although a smattering of detection methods \cite{shi2018image,bappy2019hybrid} have considered the specific contents of images, they do not fully explore the cooperation of the specific image contents with the image fingerprints learned by CNNs, which partially explains their non-promising detection results.

In summary, the existing CNN-based detection methods have not yet achieved the expected performance and generally need a pre-training process or training on a large number of training samples as listed in Table \ref{tab2}. The low performance of these methods can be attributed to two aspects that have not been fully considered: the intrinsic characteristic of image splicing forgery and the learning of image fingerprints. We propose a novel dual-encoder network (D-Net) for splicing forgery detection to resolve these issues. Our main contributions are summarized below.
\begin{table}[htbp!]
	\setlength{\abovecaptionskip}{0.1 cm}
	\setlength{\belowcaptionskip}{6 pt}
	\renewcommand\arraystretch{1.5}
	\caption{Numbers of Samples in the Training/Pre-training Processes of Existing CNN-Based Detection Methods for Image Splicing Forgery.}
	\begin{center}
		\setlength{\tabcolsep}{8pt}{
			\arrayrulecolor{black}
			\begin{tabular}{|c|c|c|}
				\arrayrulecolor{black}\hline
				 \textbf{Method} &  \textbf{Pre-training} &  \textbf{Training set} \\
				\hline\hline
				Rao, \emph{et al}. \cite{rao2016deep}	&No	&16K(patch)	\\ \hline
				Zhang, \emph{et al}. \cite{zhang2016image}	&No	&100K(patch)	\\ \hline
				Bappy, \emph{et al}. \cite{bappy2017exploiting}	&No	&100K(patch)	\\ \hline
				Wei, \emph{et al}. \cite{wei2018c2r}	&No	&335K(patch)	\\ \hline
				Bayar, \emph{et al}. \cite{bayar2016deep}	&No &600K(patch)  \\ \hline
				Bayar, \emph{et al}. \cite{bayar2018constrained}	&No	&600K(patch)	\\ \hline
				Bi, \emph{et al}. \cite{bi2019rru}	&No	&4K(image)	\\ \hline
				Shi, \emph{et al}. \cite{shi2018image}	&No	&50K(patch)	\\ \hline
				Salloum, \emph{et al}. \cite{salloum2018image}	&No &5K(image)  \\ \hline
				Xiao, \emph{et al}. \cite{xiao2020image}	&No	&335K(patch)	\\ \hline
				Liu, \emph{et al}. \cite{liu2020exposing}	&No	&160K(patch)	\\ \hline
				Cozzolino, \emph{et al}. \cite{cozzolino2019noiseprint}&No	&10K(patch)	\\ \hline
				Huh, \emph{et al}. \cite{huh2018fighting}&No	&4000K(patch)	\\ \hline
				Marra, \emph{et al}. \cite{marra2020full}&No	& $\infty$(image)	\\ \hline
				Zhou, \emph{et al}. \cite{zhou2018learning}	&Yes&42K(image)	\\ \hline
				Bappy, \emph{et al}. \cite{bappy2019hybrid}	&Yes &65K(image)  \\ \hline
				Wu, \emph{et al}. \cite{wu2019mantra}	&Yes &1.25M(patch)	\\ \hline
				Ours                                    & No &1K(image) \\
				\hline
		\end{tabular}}
	\end{center}
	\label{tab2}
\end{table}
\begin{itemize}
	\item[$\bullet$] Global perspective: we regard the image splicing forgery detection as a pixel-level binary classification task that distinguishes the tampered and non-tampered regions in splicing forged images and whether a pixel is forged or not depends on the neighbored pixels and the global information. Based on the end-to-end D-Net, we constructed a spatial pyramid global-feature extraction module (SPGFE) that expands the global insight further, achieving better detection performance than the existing methods.
	\item[$\bullet$] Dual-encoder: we employed an unfixed encoder that autonomously learns the image fingerprints and a fixed encoder that intentionally provides the prior structural information of the pixels to each stage of the detection network. The image fingerprints and pixel structures cooperate to enhance the performance of the learning and inference of the detection network.
	\item[$\bullet$] Reduced number of training samples: Owing to its structure, D-Net gains insight into the image splicing forgery from local and global perspectives. Moreover, it learns forensic features rather than semantic features, thus showing good generalization ability. Therefore, it achieves promising performance without requiring a large number of training samples and a pre-training step.
\end{itemize}

The remainder of this paper is organized as follows. Section \ref{2} introduces the existing CNN-based detection methods for image forgery detection. Section \ref{3} briefly presents the proposed D-Net, and then introduces the dual-encoder structure and SPGFE module. Section \ref{4} assesses the detection performances of the proposed D-Net and (for comparison) the state-of-the-art detection methods. The paper concludes with Section \ref{5}.

\section{Related Work}\label{2}
As summarized in Table \ref{tab1}, some proposed detection networks only discriminate whether an image is original or forged. Rao, \emph{et al}. \cite{rao2016deep} replaced the first layer of the VGG network with a steganalysis rich model (SRM) \cite{fridrich2012rich} to obtain the image noise fingerprint of the input image. Whether the image has been forged or not is then assessed by the VGG network. Bayar, \emph{et al}. \cite{bayar2016deep,bayar2018constrained} designed a constrained convolution layer that suppresses the image contents and adaptively learns the forgery features. After training on ten different datasets, their network detects multiple types of global image forgery. Zhou, \emph{et al}. \cite{zhou2018learning} introduce Faster Region-CNN with an SRM layer that extracts the image noise fingerprint, providing additional evidence for the classiﬁcation of multiple types of image local forgery. Marra, \emph{et al}. \cite{marra2020full} propose an end-to-end-trainable framework for image forgery detection, comprising extraction, aggregation, and classification blocks. They used a gradient checkpoint strategy to solve the memory problem and perform end-to-end training with full-size and full-resolution images.

Other detection networks focus on locating the tampered regions. Zhang, \emph{et al}. \cite{zhang2016image} utilized a three-layer CNN with a multi-layer perceptron (MLP). Wei, \emph{et al}. \cite{wei2018c2r} propose a two-stage VGG network, and Xiao, \emph{et al}. \cite{xiao2020image} designed a two-stage network that locates the tampered regions on both coarse and refined scales and generates the final detected tampered regions by an adaptive clustering approach. Cozzolino, \emph{et al}. \cite{cozzolino2019noiseprint} used the Siamese network to extract the noiseprint of the camera and use it to locate the tampered regions. These four methods \cite{zhang2016image,wei2018c2r,xiao2020image,cozzolino2019noiseprint} were performed on image patches, which causes high computational complexity. Moreover, their detection results are either inaccurate or require complex post-processing. Salloum, \emph{et al}. \cite{salloum2018image}  propose a multi-task FCN (MFCN) with two output branches for multi-task learning. One branch learns the tampered regions, while the other branch learns the boundaries of the tampered regions. Bi, \emph{et al}. \cite{bi2019rru} present a ringed residual U-Net (RRU-Net), in which the ringed-residual structure is formed from residual-propagation and residual-feedback modules. The RRU-Net structure enhances feature propagation and encourages feature reuse. Both MFCN and RRU-Net \cite{salloum2018image,bi2019rru} are difficult to train, and the detection results could be further improved. Bappy, \emph{et al}. \cite{bappy2017exploiting} inserted a long short-term memory (LSTM) network between the second and third layers of a five-layer CNN for learning the spatial correlations among the non-overlapping image patches. Shi, \emph{et al}. \cite{shi2018image} extracted the image noise fingerprint by an SRM layer, then employed a VGG network that learns the differences between the tampered and non-tampered regions. Finally, the tampered regions are detected with the help of seven statistical features calculated in the frequency domain based on \cite{zhang2016image}. Huh, \emph{et al}. \cite{huh2018fighting} used the automatically recorded photo EXIF metadata as supervision labels, and then used the Siamese network to determine whether its content could have been produced by a single imaging pipeline. Liu, \emph{et al}. \cite{liu2020exposing} extracted the noise and JPEG compression features by DenseNet, and hence located the tampered regions. However, these four methods \cite{bappy2017exploiting,shi2018image,huh2018fighting,liu2020exposing} learn only a particular image fingerprint, and easily obtain false decisions. Bappy, \emph{et al}. \cite{bappy2019hybrid} extracted the resampling features by a Laplacian filter and LSTM, and obtained the spatial domain features by an encoder-decoder network. The resampling and spatial domain features were then fused to locate the tampered regions. Wu, \emph{et al}. \cite{wu2019mantra} present an image manipulation trace-feature extractor and a local anomaly detection network. Zhang, \emph{et al}. \cite{zhang2020dense} used steganalysis rich model (SRM) to obtain the noise residual of the image, and then used DenseNet for binary classification and finally obtained the final prediction result through upsampling. The methods of Bappy, \emph{et al}. \cite{bappy2019hybrid}, Wu, \emph{et al}. \cite{wu2019mantra} and Zhang, \emph{et al}. \cite{zhang2020dense} attempt to strengthen the autonomous learning of image fingerprints by a particular image fingerprint, but the pre-training requires many forged samples to improve the detection results.


\section{The Proposed Dual-Encoder Network}\label{3}

\begin{figure*}[htbp!]
	\centering
	\includegraphics[width=0.8\linewidth]{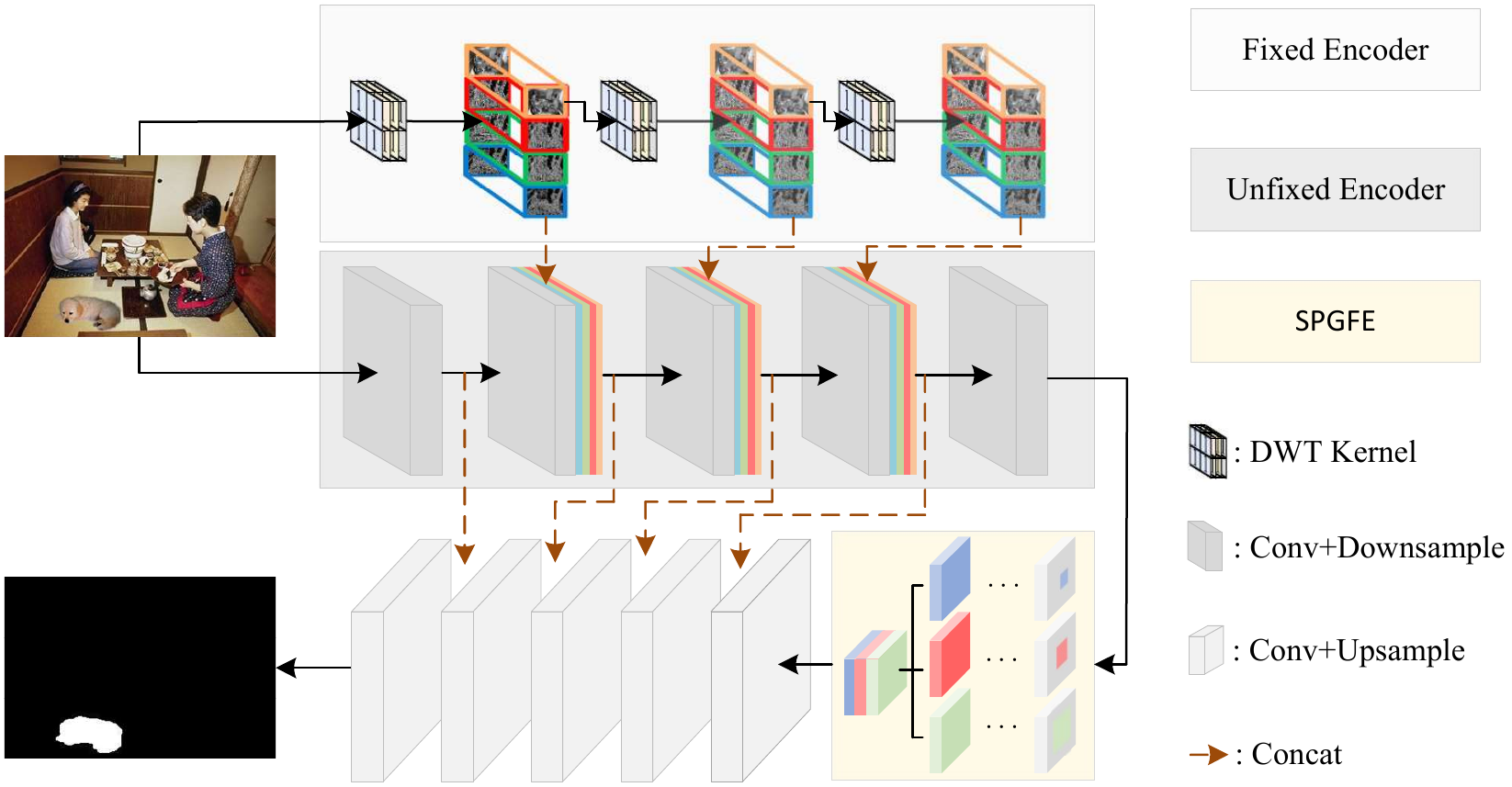}
	\caption{Structure of the proposed D-Net.}
	\label{fig2}
\end{figure*}

Our proposed dual-encoder network (D-Net) (as shown in Fig. \ref{fig2}) locates the tampered regions in splicing forgery images. Utilizing the powerful feature learning and mapping abilities of CNNs, the unfixed encoder of D-Net autonomously learns the image fingerprints that distinguish the tampered and non-tampered regions. The fixed encoder extracts the direction information of the pixels in the image and cooperates with the unfixed encoder during the learning and decision-making of the detection network. Next, the global insight of D-Net is expanded by an SPGFE, which obtains low-resolution feature maps with global features. Finally, the decoder learns the mapping from the low-resolution feature maps and makes pixel-wise predictions of the tampered regions. Owing to its structure, D-Net gains insights into the image splicing forgery from a global perspective; accordingly, it maintains steady performance when detecting different cases of splicing forgeries without requiring a large number of training samples or a pre-training process. We describe the unfixed encoder, fixed encoder, and SPGFE module of the proposed D-Net in detail in the subsequent subsections.
\subsection{The Unfixed Encoder in Dual-encoder of D-Net}\label{3.1}
To effectively utilize the feature learning and mapping ability of CNNs in the CNN-based detection method for image splicing forgery, we must analyze the intrinsic characteristics of image splicing forgery in advance. Based on the similarity of the tampered regions to their surrounding regions, splicing forgery can be generally divided into object splicing forgery and smooth splicing forgery. If the tampered and surrounding regions are similar, the boundaries of the tampered regions are indistinct. Methods that distinguish between tampered and non-tampered regions mainly detect the differences in the image fingerprints of both regions. If the tampered regions are objects, their boundaries will introduce real edges in splicing forgery images. In this case, the direction information of the pixels is an important addition to the image fingerprints. Because the image fingerprints are necessary in both cases of splicing forgery, learning the image fingerprints is the foremost task of forgery-detection networks.

When learning the image fingerprints that distinguish tampered from non-tampered regions, the network depth must be carefully chosen because the image fingerprints generally possess basic image properties rather than high-level semantic features. A deep network learns the higher-level semantic information and loses the basic image properties. Conversely, the learned features of a shallow network are insufficient for exploring the different image fingerprints of the tampered and non-tampered regions. The operation of down-sampling in the detection network should also be carefully noted. The forgery prediction of the decoder is based on the low-resolution feature maps that contain the differences between the image fingerprints. To distinguish between the tampered and non-tampered regions, the image fingerprints must be maintained and even enhanced. Down-sampling by max-pooling will only keep the element with the largest value in the local window. As the image fingerprints are a low-level feature, they are generally not the largest element in the local window, so using the max-pooling will reduce or miss the image fingerprints.


As the basic network in the proposed feature pyramid network (FPN) \cite{lin2017feature}, FPN delivers promising performance on object detection. To analyze the effect of the network depth and down-sampling operation, we varied the depth of FPN and performed down-sampling by a convolution of stride 2 or max-pooling. The experiments are carried out on two public datasets, CASIA \cite{dong2013casia} and COLUMB\cite{hsu2006detecting}. The CASIA dataset \cite{dong2013casia} includes 750 splicing forgery images for training and validation and 100 splicing forgery images for testing. The COLUMB dataset \cite{hsu2006detecting} includes 135 splicing forgery images for training and validation and 45 splicing forgery images for testing. The FPN with max-pooling and stride-2 convolution down-sampling are denoted by FPN-N-M, and FPN-N-C, respectively. The number N in the middle part of the name represents the number of down-sampling operations in the encoder part of FPN.
\begin{figure}[htbp!]
	\begin{center}
		\includegraphics[width=0.95\linewidth]{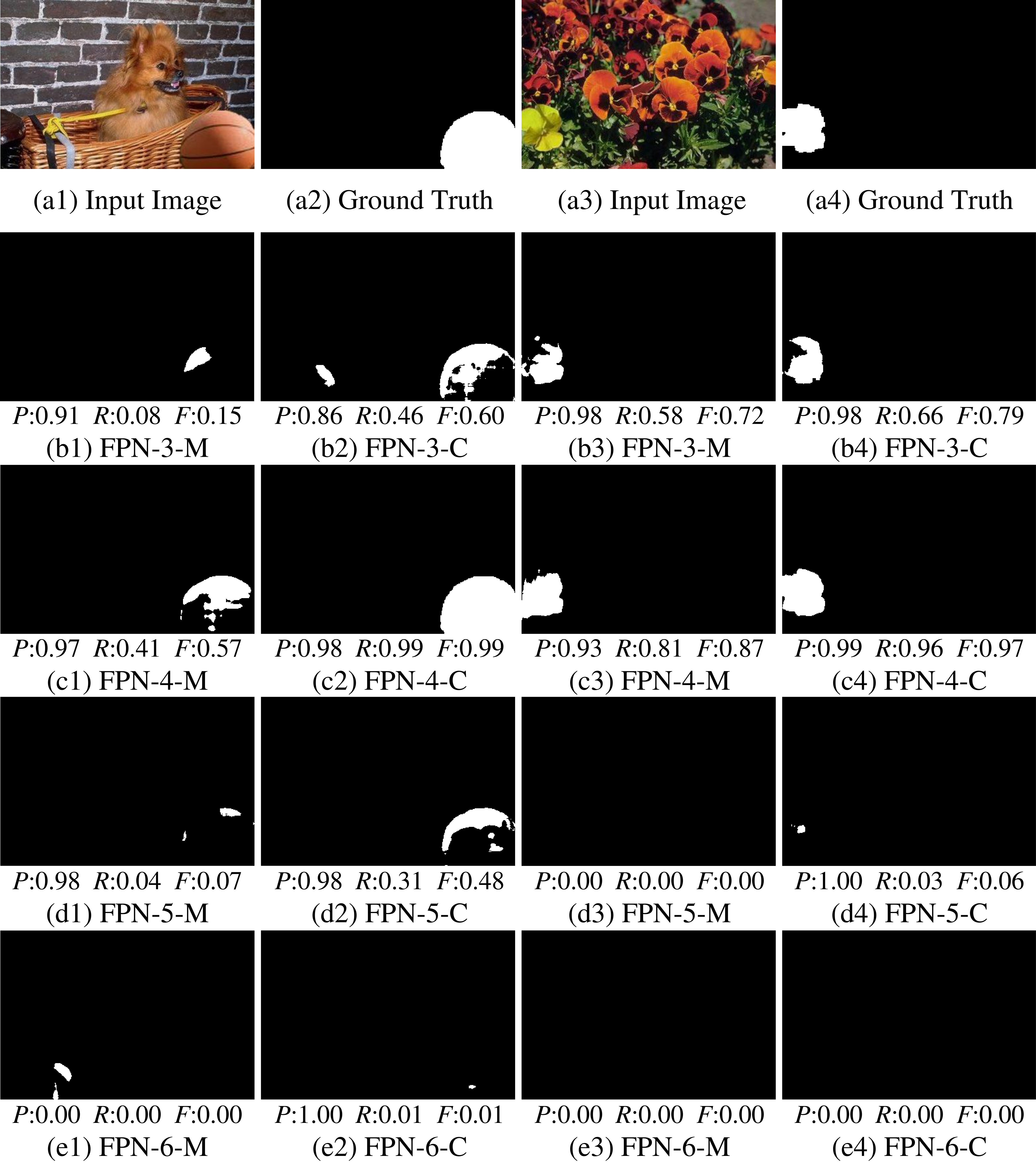}
	\end{center}
	\caption{Regions of two splicing forgery images, detected by FPNs with different depths and down-sampling operations. Shown are the splicing forgery images and their ground-truth masks (row \emph{a}) and the detection results of FPN-3, FPN-4, FPN-5, and FPN-6 (row \emph{b}, \emph{c}, \emph{d}, and \emph{e}, respectively). Column \emph{a}1 and \emph{a}3 are the detection results of max-pooling, and column \emph{a}2 and \emph{a}4 are the detection results of stride-2 convolution. }
	\label{fig3}
\end{figure}
For a subjective analysis, we selected two splicing forgery images from the testing sets and evaluated their results in different cases of FPN. As shown in Fig. \ref{fig3}, regardless of whether the tampered regions are similar to their non-tampered surroundings (Fig. \ref{fig3}-(a1)), or are objects inserted into the image (Fig. \ref{fig3}-(a3)), FPN with depth 4 achieves the best detection result on both cases. Moreover, FPNs of the same depth consistently achieve higher performance when down-sampled by a convolution of stride 2 than when down-sampled by max-pooling. For objective evaluation, the detection results of the FPN variants on both datasets are compared in Table \ref{tab3}. Again, FPN-4-C still shows the best performance. The experimental results verify the importance of selecting a suitable network depth and down-sampling operation. In subsequent analyses, we employ FPN-4-C as the unfixed encoder and decoder in the proposed D-Net.
\begin{table}[h]
	\setlength{\abovecaptionskip}{0.1 cm}
	\setlength{\belowcaptionskip}{6 pt}
	\caption{Detection Results of FPNs with Different Network Depths and Down-Sampling Operations on the CASIA and COLUMB Datasets.}
	\begin{center}
		\renewcommand\arraystretch{1.5}
		\scalebox{0.85}{
			\arrayrulecolor{black}
			\begin{tabular}{|c|m{1cm}<{\centering}|m{0.8cm}<{\centering}|m{0.8cm}<{\centering}|m{1cm}<{\centering}|m{0.8cm}<{\centering}|m{0.8cm}<{\centering}|}
				\arrayrulecolor{black}\hline
				\multicolumn{1}{|c|}{}                         & \multicolumn{6}{c|}{\small \textbf{Detection Result}}                                                         \\ \arrayrulecolor{black}\cline{2-7}
				\multicolumn{1}{|c|}{}                         & \multicolumn{3}{c|}{CASIA}    & \multicolumn{3}{c|}{COLUMB}           \\ \arrayrulecolor{black}\cline{2-7}
				\multicolumn{1}{|c|}{\multirow{-3}{*}{\small \textbf{Method}}} & \textit{Precision} & \textit{Recall} & \textit{F}     & \textit{Precision}         & \textit{Recall} & \textit{F}     \\ \hline \hline
				FPN-3-M                                                               & 0.795              & 0.736           & 0.765          & 0.832                      & 0.43            & 0.567          \\ \hline
				FPN-4-M                                                               & \textbf{0.815}     & 0.735           & 0.773          & 0.906                      & 0.535           & 0.661          \\ \hline
				FPN-5-M                                                               & 0.794              & 0.722           & 0.756          & 0.853                      & 0.502           & 0.632          \\ \hline
				FPN-6-M                                                               & 0.729              & 0.639           & 0.681          & \multicolumn{1}{c|}{0.849} & 0.487           & 0.619          \\ \hline \hline
				FPN-3-C                                                               & 0.801              & 0.76            & 0.78           & 0.835                      & 0.531           & 0.649          \\ \hline
				FPN-4-C                                                               & 0.813              & 0.764           & \textbf{0.788} & \textbf{0.923}             & \textbf{0.547}  & \textbf{0.686} \\ \hline
				FPN-5-C                                                               & 0.774              & \textbf{0.782}  & 0.778          & 0.866                      & 0.538           & 0.664          \\ \hline
				FPN-6-C                                                               & 0.752              & 0.648           & 0.696          & 0.848                      & 0.525           & 0.649          \\ \hline
		\end{tabular}}
	\end{center}
	\label{tab3}
\end{table}
\subsection{The Fixed Encoder in Dual-encoder of D-Net}\label{3.2}

The unfixed encoder autonomously learns the image fingerprints that differentiate between the tampered and non-tampered regions, as shown in the experimental results of FPN-4-C in subsection \ref{3.1}. However, in some state-of-the-art CNN-based detection methods for image splicing forgery, the detection networks \cite{rao2016deep,bappy2017exploiting,shi2018image,huh2018fighting,liu2020exposing} only learn the particular image fingerprint, whereas others \cite{zhou2018learning,bappy2019hybrid,wu2019mantra,zhang2020dense,marra2020full} attempt to strengthen the autonomous learning of networks by learning a particular image fingerprint. The particular image fingerprint, which generally exists in images, is thought to directly improve the learning effectiveness of the detection network.

\begin{figure}[tbp!]
	\begin{center}
		\includegraphics[width=1.0\linewidth]{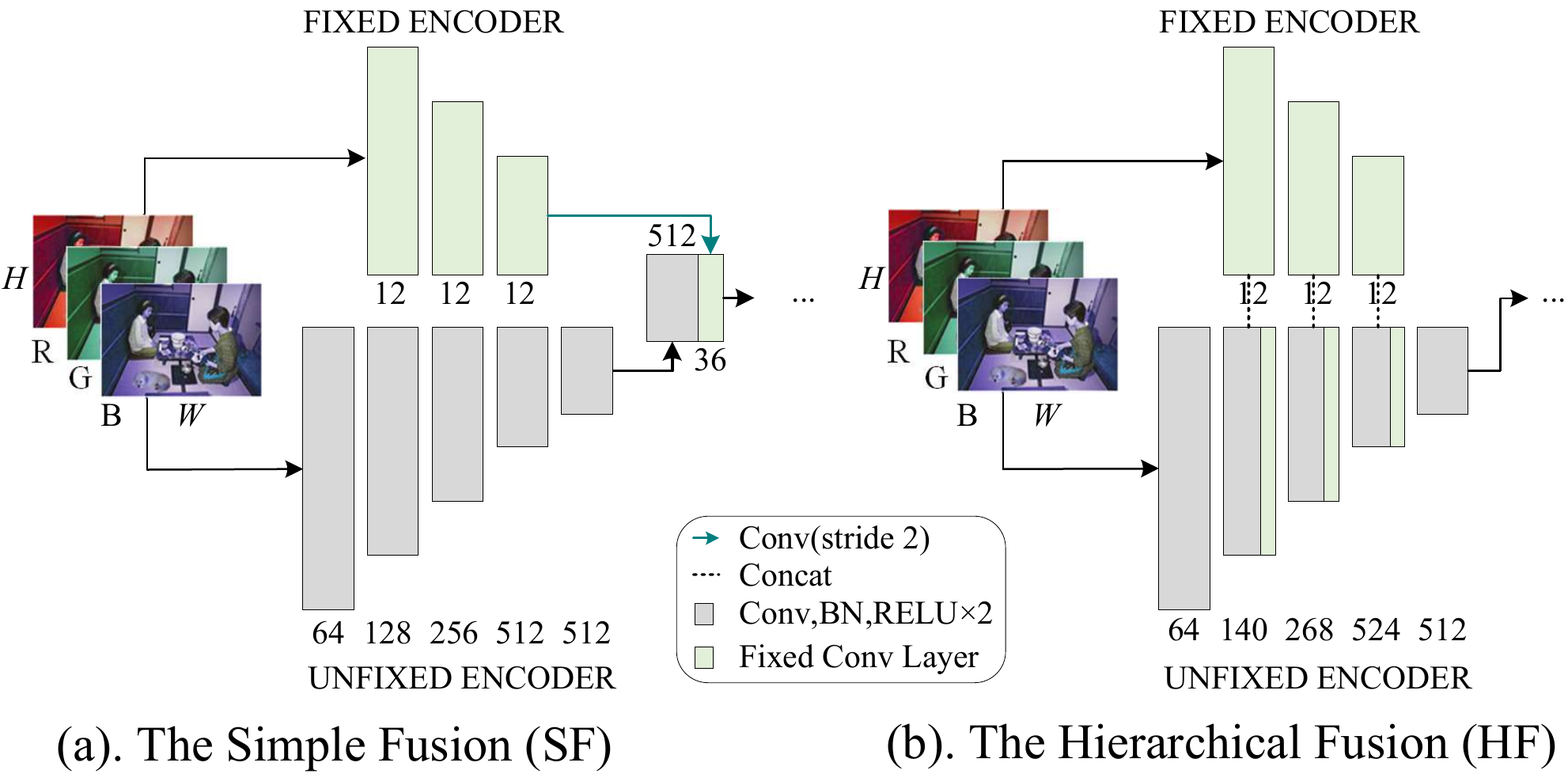}
	\end{center}
	\caption{ Structures of the two concatenation methods: (a) simple fusion and (b) hierarchical fusion.}
	\label{fig6}
\end{figure}

For this issue, we construct a fixed SRM encoder with three SRM kernels. The SRM has proven effective in existing detection methods for image splicing forgery \cite{rao2016deep,shi2018image,zhou2018learning,wu2019mantra,zhang2020dense}. The fixed SRM encoder consists of an SRM layer and three convolutional layers. The SRM layer with the three SRM kernels obtains a noisy feature map through three channels from an RGB image. The noisy map is passed through three convolution layers, each with a kernel size of 5×5, a convolution stride of 2. Finally, the output feature of the fixed SRM encoder is down-sampled and simply fused with the output feature of the unfixed encoder. This fusion method is shown in Fig. \ref{fig6}-(a) and denoted as the simple fusion method (SF).



We experiment on FPN-4-C and the fixed SRM encoder. The datasets and setting of parameters are the same as subsection \ref{3.1}. The experimental results are listed in Table \ref{tab5}(the first two rows). For the COLUMB dataset, the \emph{F} score is improved by additionally providing the image noise fingerprint to FPN-4-C. For the CASIA dataset, the improvement was not obvious. This may be due to the tampered region of the COLUMB dataset is large and meaningless, and the image noise fingerprint of the tampered region and the non-tampered region is quite different, while the tampered region of the CASIA dataset is relatively samll, and the image noise fingerprint of the tampered region and the non-tampered region is not obvious. Therefore, providing the image noise fingerprint maybe not a good way. How to effectively learn image fingerprints should be investigated more deeply according to the intrinsic characteristic of image splicing forgery.

\begin{figure}[htbp!]
	\begin{center}
		\includegraphics[width=1.0\linewidth]{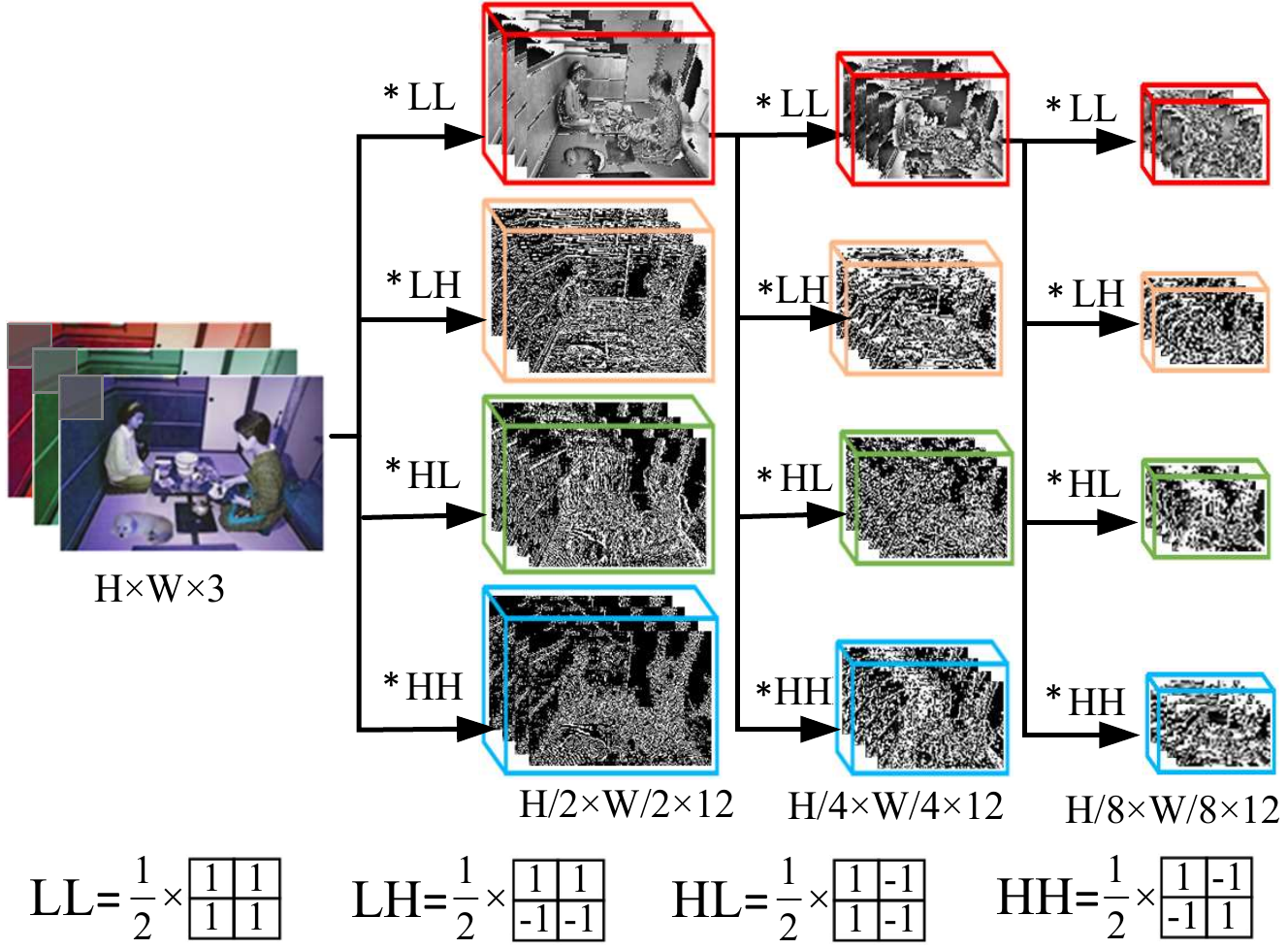}
	\end{center}
	\caption{Structure of the fixed DWT encoder.}
	\label{fig5}
\end{figure}

\begin{table}[]
	\setlength{\abovecaptionskip}{0.1 cm}
	\setlength{\belowcaptionskip}{6 pt}
	\caption{Detection Results of Different Concatenation Methods on the CASIA and COLUMB Datasets.}
	\begin{center}
		\renewcommand\arraystretch{1.5}
		\setlength{\tabcolsep}{3pt}{
			\scalebox{0.82}{
				\begin{tabular}{|m{3cm}<{\centering}|m{1.1cm}<{\centering}|m{0.8cm}<{\centering}|m{0.8cm}<{\centering}|m{1.1cm}<{\centering}|m{0.8cm}<{\centering}|m{0.8cm}<{\centering}|}
					\hline
					\multirow{3}{*}{ \textbf{Method}}                                                                 & \multicolumn{6}{c|}{\textbf{Detection Result}}                                                                                                               \\ \cline{2-7}
					& \multicolumn{3}{c|}{CASIA}                                               & \multicolumn{3}{c|}{COLUMB}                                              \\ \cline{2-7}
					& \emph{Precision}              & \emph{Recall}                 & \emph{F}                      & \emph{Precision}              & \emph{Recall}                 & \emph{F}                      \\  \hline \hline
					FPN-4-C                                                                                       & 0.813                           & 0.764                           & 0.788                           & 0.923                           & 0.547                           & 0.686                           \\ \hline
			\end{tabular}}
			
			\renewcommand\arraystretch{1.1}
			\scalebox{0.82}{
				\begin{tabular}{|m{3cm}<{\centering}|m{1.1cm}<{\centering}|m{0.8cm}<{\centering}|m{0.8cm}<{\centering}|m{1.1cm}<{\centering}|m{0.8cm}<{\centering}|m{0.8cm}<{\centering}|}
					\multirow{3}{*}{\begin{tabular}[c]{@{}c@{}}FPN-4-C\\ SF\\ The fixed SRM encoder\end{tabular}}  & \multirow{3}{*}{0.807}          & \multirow{3}{*}{0.775}          & \multirow{3}{*}{0.791}          & \multirow{3}{*}{0.819}          & \multirow{3}{*}{0.646}          & \multirow{3}{*}{0.722}          \\
					&                                 &                                 &                                 &                                 &                                 &                                 \\
					&                                 &                                 &                                 &                                 &                                 &                                 \\ \hline
					\multirow{3}{*}{\begin{tabular}[c]{@{}c@{}}FPN-4-C\\ HF\\ The fixed SRM encoder\end{tabular}}  & \multirow{3}{*}{\textbf{0.835}} & \multirow{3}{*}{0.778}          & \multirow{3}{*}{0.805}          & \multirow{3}{*}{0.918}          & \multirow{3}{*}{0.629}          & \multirow{3}{*}{0.747}          \\
					&                                 &                                 &                                 &                                 &                                 &                                 \\
					&                                 &                                 &                                 &                                 &                                 &                                 \\ \hline
					\multirow{3}{*}{\begin{tabular}[c]{@{}c@{}}FPN-4-C \\ SF\\ The fixed DWT encoder\end{tabular}} & \multirow{3}{*}{0.824}          & \multirow{3}{*}{0.782}          & \multirow{3}{*}{0.803}          & \multirow{3}{*}{0.906}          & \multirow{3}{*}{0.651}          & \multirow{3}{*}{0.757}          \\
					&                                 &                                 &                                 &                                 &                                 &                                 \\
					&                                 &                                 &                                 &                                 &                                 &                                 \\ \hline
					\multirow{3}{*}{\begin{tabular}[c]{@{}c@{}}FPN-4-C \\ HF\\ The fixed DWT encoder\end{tabular}} & \multirow{3}{*}{0.833}          & \multirow{3}{*}{\textbf{0.794}} & \multirow{3}{*}{\textbf{0.814}} & \multirow{3}{*}{\textbf{0.926}} & \multirow{3}{*}{\textbf{0.648}} & \multirow{3}{*}{\textbf{0.763}} \\
					&                                 &                                 &                                 &                                 &                                 &                                 \\
					&                                 &                                 &                                 &                                 &                                 &                                 \\ \hline
			\end{tabular}}
		}
	\end{center}
	\label{tab5}
\end{table}

In most splicing forgery images, the tampered regions are objects with boundaries that introduce real edges. In this case, in addition to image fingerprints, the direction information of the pixels should be supplemented since it represents the boundaries of the tampered regions. However, the direction information of the pixels is difficult to retain through CNN's processing. The direction information of pixels is additionally provided to the detection network, which may assist the detection network.

To test this idea, we perform a three-level Haar discrete wavelet transform (DWT) on each color channel of the input RGB image, which is shown in Fig. \ref{fig5}. Thirty-six (3×12) feature maps are obtained by the three-level wavelet decomposition (12 (3×4) feature maps at each level). The output feature of the fixed DWT encoder is fused with the output feature maps of the unfixed encoder by the simple fusion method (SF), as shown in Fig. \ref{fig6}-(a). The fixed DWT encoder was experimentally evaluated under the same conditions as the fixed SRM encoder, and the detection results are listed in Table \ref{tab5}. All experimental results are improved from those of the fixed SRM encoder, confirming that the prior structural information of the pixels provided by the fixed DWT encoder assists the FPN-4-C in detecting and depicting the tampered regions.

The above results confirm the effectiveness of providing the prior structural of the pixels to FPN-4-C in the decoding process, but whether the prior structural information of the pixels can also assist FPN-4-C in the encoding process. For testing this idea, we propose a hierarchical fusion (HF) method between the fixed and unfixed encoder, which concatenates the output feature of each level in the fixed encoder to the output feature of the corresponding level in the unfixed encoder as shown in Fig. \ref{fig6}-(b). FPN-4-C is experimentally evaluated for different fixed encoders and concatenation methods (the datasets and setting of parameters are the same as subsection \ref{3.1}). The detection results on the CASIA and COLUMB datasets are listed in Table \ref{tab5}. In the FPN-4-C networks with the fixed SRM and DWT encoders, the HF improves the image-splicing forgery detection on both datasets (from that of SF). The prior structural information of the pixels provided by the fixed DWT encoder not only benefits the prediction of the detection network but also indicates the detection network where image fingerprints may be different in the encoding process.

\begin{figure}[htbp!]
	\begin{center}
		\includegraphics[width=1.0\linewidth]{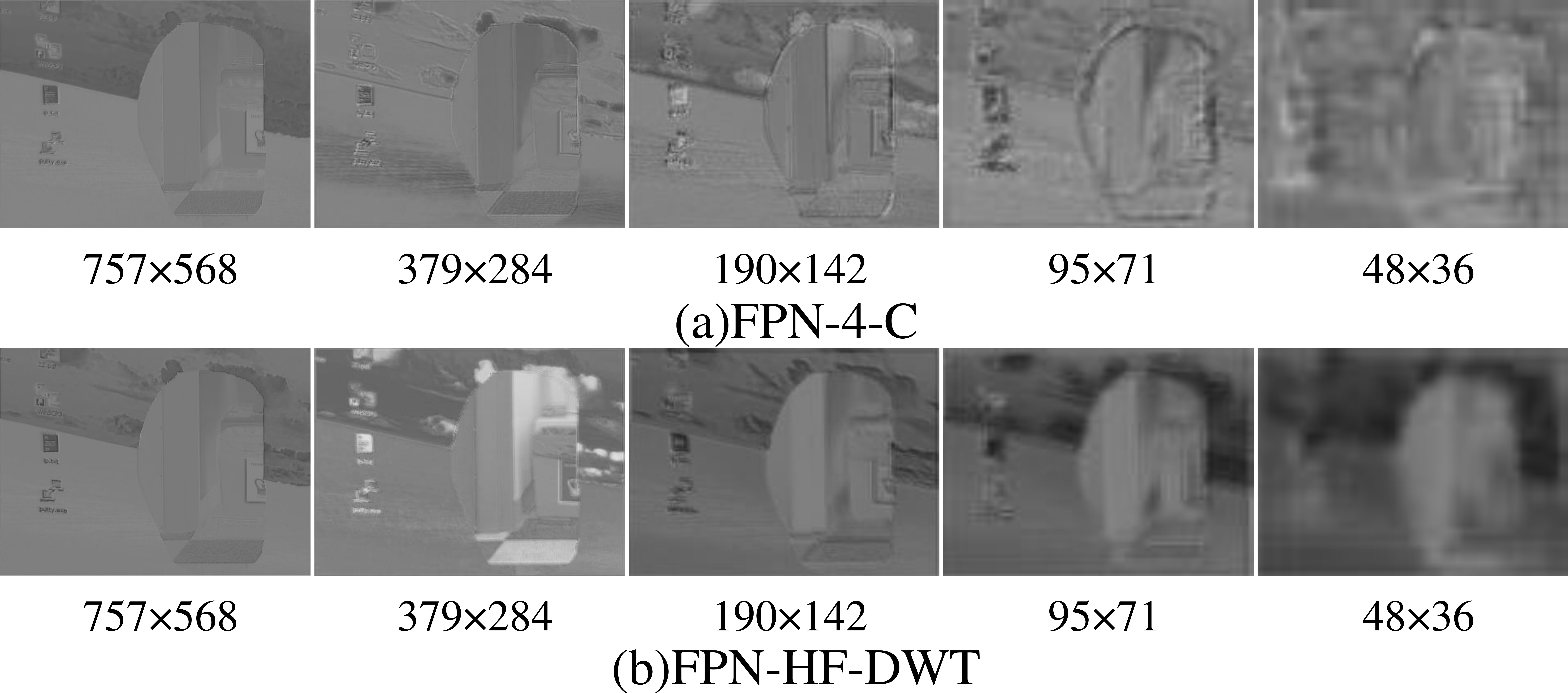}
	\end{center}
	\caption{The output feature maps of each layer of FPN-4-C and FPN-HF-DWT encoders. The forgery image is canong3\_canonxt\_sub\_28.tif of the COLUMB dataset. Depicted are the output feature maps of each layer of the FPN-4-C encoder (row \emph{a}) and the output feature maps of each layer of the FPN-HF-DWT encoder (row \emph{b}). The $1^{st}$ to $5^{th}$ columns are the output feature maps of the first to fifth layer respectively. The bottom of each column of images is the actual resolution of each layer of feature maps.}
	\label{fig7-a}
\end{figure}

To show the effectiveness of providing the direction information of the pixels in FPN-4-C more intuitively, we have visualized the output feature maps of each layer of FPN-4-C and FPN-HF-DWT encoders. As shown in Fig. \ref{fig7-a}, the output feature maps of the first layer ($1^{st}$ column in Fig. \ref{fig7-a}) of the two encoders are roughly the same, but after providing the prior structural information of the pixels in FPN-4-C ($2^{nd}$ to $4^{th}$ columns in Fig. \ref{fig7-a}), 

FPN-HF -DWT is obviously more inclined than FPN-4-C to learn the image fingerprints that differentiate between the tampered and non-tampered regions. Therefore, the output feature maps of FPN-HF-DWT on the fifth layer ($5^{th}$ column in Fig. \ref{fig7-a}) are much more accurate than that of FPN-4-C.

\subsection{Spatial Pyramid Global-Feature Extraction Module}\label{3.3}

\begin{figure*}[htbp!]
	\begin{center}
		\includegraphics[width=0.85\linewidth]{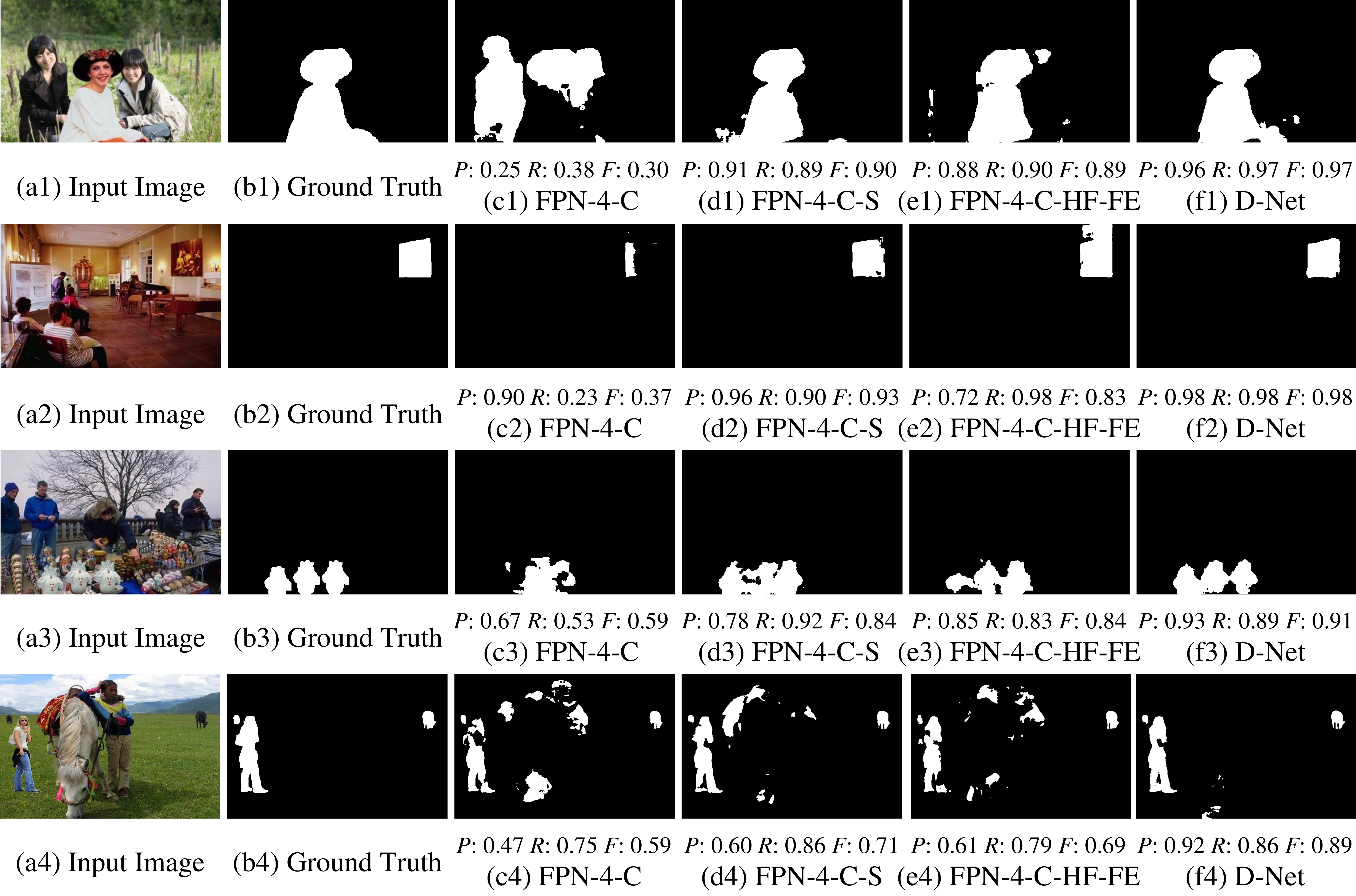}
	\end{center}
	\caption{Regions of five splicing forgery images, detected by FPN-4-C with and without the fixed DWT encoder and SPGFE. The $1^{st}$, $2^{nd}$ and $3^{rd}$ rows contain a tampered regions on different scales, respectively. The $4^{th}$ row contains multiple tampered regions on different scales. Column \emph{a} is the forgery image; column \emph{b} is the ground-truth; column \emph{c} is the detection result of FPN-4-C; column \emph{d} is the detection result of FPN-4-C with SPGFE; column \emph{e} is the detection result of FPN-4-C with the fixed DWT encoder (dual-encoder structure); column \emph{f} is the detection result of dual-encoder network structure with SPGFE (D-Net). }
	\label{fig9}
\end{figure*}

In image splicing forgery detection, the tampered and non-tampered regions should be distinguished by a global binary classification. As shown in the fourth row of Fig. \ref{fig9}, one splicing forgery image contains multiple tampered regions on different scales. Therefore, the decision of the detection network must cross local regions and local objects. However, like the existing end-to-end detection networks \cite{bayar2016deep,bayar2018constrained,salloum2018image,bi2019rru,zhou2018learning,bappy2019hybrid,wu2019mantra,zhang2020dense,marra2020full}, the above network structure focuses only on local objects, which is insufficient for accurate detection of splicing forgery. For expanding the global perspective of the detection network further, we design a spatial pyramid global-feature extraction module (SPGFE) after the dual-encoder.

\begin{figure}[h]
	\begin{center}
		\includegraphics[width=1.0\linewidth]{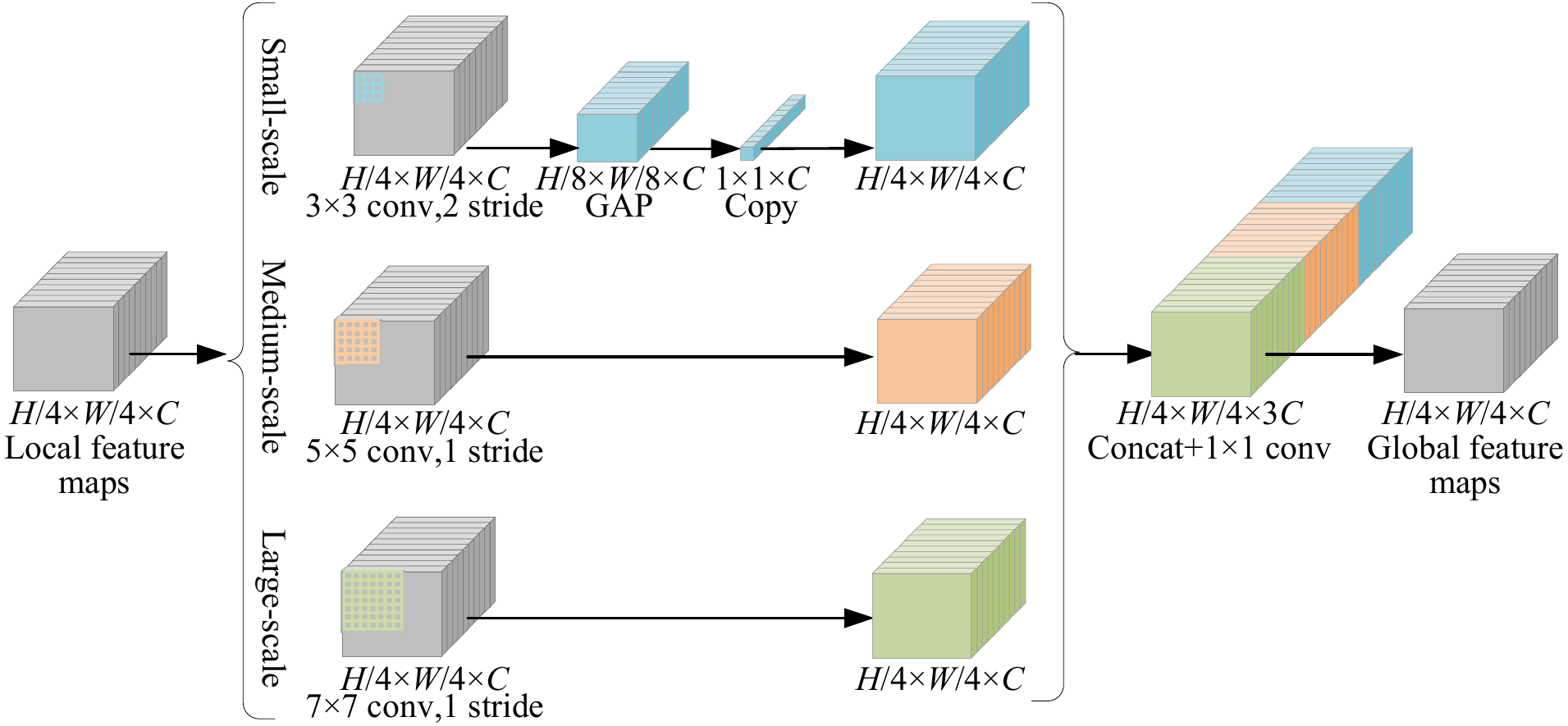}
	\end{center}
	\caption{Structure of the spatial pyramid global-feature extraction module.}
	\label{fig8}
\end{figure}

The SPGFE is inspired by human visual decision-making. To observe a local part of an object, humans approach the object; to obtain a global view, they step away from the object. The proposed SPGFE module is shown in Fig. \ref{fig8}. 
The local feature maps generated by the fixed and unfixed encoders pass through three branches that generate global feature maps on different scales. The small-scale, medium-scale and large-scale global feature maps are concatenated and operated through a 1×1 convolution layer, which outputs the final multi-scale global feature maps. The three convolution kernels with different sizes are equivalent to different viewing distances from an object.

The SPGFE module is experimentally evaluated on the dataset described in subsection \ref{3.1}, under the same parameter settings. Various FPN-4-C configurations are tested: FPN-4-C combined with SPGFE (FPN-4-C-S), the fixed DWT encoder combined with FPN-4-C by HF (FPN-4-C-HF-FE), and FPN-4-C-HF-FE combined with SPGFE to expand the global insight (FPN-4-C-HF-FE-S). For a subjective analysis, four splicing forgery images are selected from the testing sets, and the detection results are shown in Fig. \ref{fig9}. The SPGFE clearly reduces the false detection rate and locates tampered regions of different sizes. As an objective evaluation, the detection results on the two datasets are listed in Table \ref{tab6}. The SPGFE undoubtedly improves the overall performances of the FPN-4-C and FPN-4-C-HF-FE detection networks. After the complete set of analyses in this section, the performance of the proposed framework is optimized by cooperating with the fixed DWT encoder with FPN-4-C, fusing their feature maps by HF, and processing the local features of the dual-encoder through the SPGFE module.

\begin{table}[h]


	
	\setlength{\abovecaptionskip}{1 pt}
	\setlength{\belowcaptionskip}{6 pt}
	\caption{Detection Results of the Detection Networks with and without SPGFE on the CASIA and COLUMB Datasets.}
	\begin{center}
		\renewcommand\arraystretch{1.5}
		\scalebox{0.8}{
			\setlength{\tabcolsep}{3pt}{
				\begin{tabular}{|c|m{1.1cm}<{\centering}|m{0.8cm}<{\centering}|m{0.8cm}<{\centering}|m{1.1cm}<{\centering}|m{0.8cm}<{\centering}|m{0.8cm}<{\centering}|}
					\hline
					\multicolumn{1}{|c|}{}                         & \multicolumn{6}{c|}{ \textbf{Detection Result}}                                                         \\ \cline{2-7}
					\multicolumn{1}{|c|}{}                         & \multicolumn{3}{c|}{CASIA}    & \multicolumn{3}{c|}{COLUMB}           \\ \cline{2-7}
					\multicolumn{1}{|c|}{\multirow{-3}{*}{ \textbf{Method}}} & \textit{Precision} & \textit{Recall} & \textit{F}     & \textit{Precision}         & \textit{Recall} & \textit{F}     \\ \hline \hline
					FPN-4-C                  & 0.813              & 0.764           & 0.788          & 0.923              & 0.547           & 0.686         \\ \hline
					FPN-4-C-S                & 0.833              & 0.817           & 0.825          & 0.941              & 0.701           & 0.803         \\ \hline
					FPN-4-C-HF-FE            & 0.833              & 0.794           & 0.814          & 0.926              & 0.648           & 0.763         \\ \hline
					FPN-4-C-HF-FE-S(D-Net) & \textbf{0.866}     & \textbf{0.852}  & \textbf{0.859} & \textbf{0.96}      & \textbf{0.901}  & \textbf{0.93} \\ \hline
		\end{tabular}}}
	\end{center}
	\label{tab6}
\end{table}

\section{Experiments and Discussions}\label{4}
In this section, the performance of the proposed D-Net is analyzed and evaluated in various experiments. Subsection \ref{4.1} introduces the datasets and evaluation metrics used in the experiments. Subsection \ref{4.2} describes the implementation details and training process of D-Net. Subsection \ref{4.3} introduces several existing detection methods. Finally, subsection \ref{4.4} compares the effectiveness and robustness of D-Net and several existing detection methods for image splicing forgery.

\vspace{-0.1cm}  
\subsection{Experimental datasets and evaluation metrics}\label{4.1}
The image splicing forgery detection methods will be analyzed and evaluated on four public datasets, namely, CASIA \cite{dong2013casia}, COLUMB \cite{hsu2006detecting}, NIST’16 \cite{guan2019mfc}, Fantastic Reality(FR) \cite{kniaz2019point}. The image information of each dataset is detailed in Table \ref{tab7}. The CASIA dataset contains splicing forgery, copy-move forgery, and removal forgery. The tampered regions are carefully manipulated and post-processed by methods such as filtering and blurring, but no 
\begin{table}[htbp!]
	\setlength{\abovecaptionskip}{0.1 cm}
	\setlength{\belowcaptionskip}{6 pt}
	\caption{Characteristics of the Image Splicing Forgery Datasets.}
	\begin{center}
		\renewcommand\arraystretch{1.5}
		\scalebox{0.85}{
			\setlength{\tabcolsep}{1pt}{
				\begin{tabular}{|m{1.3cm}<{\centering}|m{0.9cm}<{\centering}|m{1.5cm}<{\centering}|m{1.3cm}<{\centering}|m{1.5cm}<{\centering}|m{2.5cm}<{\centering}|}
					\hline
					\textbf{Dataset} & \textbf{Year} & \textbf{Image Format}  & \textbf{Forgery Region} & \textbf{Resolution}  & \textbf{Forged/Authentic Image} \\ \hline \hline
					CASIA       &2013 & TIFF,JPEG                & Object         & 160×240-900×600         & 5123/7491                                                         \\ \hline
					COLUMB      &2006 & TIFF                     & Smooth         & 757×568-1152×768        & 180/183                                                           \\ \hline
					NIST’16     &2016 & JPEG                     & Object         & 500×500-5616×3744       & 564/875                                                           \\ \hline
					FR          &2019 & JPEG                     & Object         & 282×800-6000×4000       & 16000/16000                                       \\ \hline
		\end{tabular}}}
	\end{center}
	\label{tab7}
\end{table}
ground-truth masks of the tampered regions are provided. The CASIA dataset includes 5123 forgery images in TIFF and JPEG format. Most of the images have a resolution of 384×256. The COLUMB dataset contains 180 forgery images in TIFF format, with typical resolutions between 757×568 and 1152×768. This dataset includes only splicing forgery, and the tampered regions are the large meaningless smooth regions. The corresponding ground-truth masks are provided. The NIST’16 dataset is a challenging dataset containing 564 forgery images in JPEG format. The three types of local forgeries-splicing, copy-move, and removal-are included. The forgery images in this dataset have been post-processed to hide any visible traces of manipulation, and their corresponding ground-truth masks are also provided. The Fantastic Reality(FR) dataset is more extensive in terms of scene variety and image count. It contains 16k tampered and 16k authentic images. This dataset only includes splicing forgery, and it also provides ground-truth masks, instance and class labels for each image.

The performances of the methods for detecting image splicing forgery are evaluated by the \emph{Precision}, \emph{Recall}, and \emph{F} rate. \emph{Precision}, defined by Eq. (\ref{eq1}), is the ratio of the correctly detected pixels to all detected pixels. \emph{Recall}, defined by Eq. (\ref{eq2}), is the ratio of correctly detected pixels to the ground-truth. \emph{TP} and \emph{FP} denote the numbers of correctly detected and erroneously detected pixels, respectively, and \emph{FN} is the number of falsely missed pixels. The \emph{F} measure is the weighted harmonic mean of the \emph{Precision} and \emph{Recall} rate, and is given by Eq. (\ref{eq3}).

\begin{equation}
\setlength{\abovedisplayskip}{3pt}
\setlength{\belowdisplayskip}{3pt}
\emph{Precision} = \frac{\emph{TP}}{\emph{TP}+\emph{FP}}
\label{eq1}
\end{equation}
\begin{equation}
\setlength{\abovedisplayskip}{3pt}
\setlength{\belowdisplayskip}{3pt}
\emph{Recall} = \frac{\emph{TP}}{\emph{TP}+\emph{FN}}
\label{eq2}
\end{equation}
\begin{equation}
\setlength{\abovedisplayskip}{3pt}
\setlength{\belowdisplayskip}{3pt}
\emph{F} = \frac{2\times\emph{Precision}\times\emph{Recall}}{\emph{Precision}+\emph{Recall}}
\label{eq3}
\end{equation}

In the following experiments,\emph{ Precision}, \emph{Recall}, and \emph{F} are averaged over all images in each case.

\subsection{Implementation details and training process of D-Net}\label{4.2}
In the pixel-wise classiﬁcation by D-Net, the network is followed by a sigmoid layer. Meanwhile, the agreement between the prediction mask $p_{i}$ and ground-truth mask $m_{i}$ corresponding to the $i^{th}$ input $x_{i}$ is measured by the cross-entropy (CE) loss. For \emph{N} samples, the cross-entropy loss is computed as:

\begin{equation}
\setlength{\abovedisplayskip}{3pt}
\setlength{\belowdisplayskip}{3pt}
L_{loss}=\frac{1}{\emph{N}} \sum_{i}||m_{i} \log \left(p_{i}\right)+\left(1-m_{i}\right) \log \left(1-p_{i}\right)||_{1}
\label{eq4}
\end{equation}

D-Net was trained using an Adam training optimizer \cite{kingma2014adam} with an initial learning rate of $3e^{-4}$, a decay rate of 0, a batch size of 16, and an epoch of 50. All training processes were implemented on a NVIDIA Tesla V100 (32G) GPU. To evaluate the actual ability of the proposed D-Net, we omited a pre-training method, and trained the network on only hundreds of forgery images.

\begin{table}[htbp!]
	\setlength{\abovecaptionskip}{1 pt}
	\setlength{\belowcaptionskip}{6 pt}
	\caption{Images used in the experimental section.}
	\begin{center}
		\renewcommand\arraystretch{1.5}
		\scalebox{0.85}{
			\setlength{\tabcolsep}{1pt}{
				\begin{tabular}
					{|c|m{1.5cm}<{\centering}|c|m{0.9cm}<{\centering}|c|c|c|c|c|}
					\hline
					\multicolumn{2}{|c|}{  \textbf{Name}}                 & \textbf{Param} & \textbf{Range}      &  \textbf{Step}  & \textbf{CASIA} & \textbf{COLUMB} &  \textbf{NIST’16} & \textbf{FR} \\ \hline \hline
					\multicolumn{2}{|c|}{Training}             & -     & -          & -     & 1350   & 125    & 184   &10800  \\ \hline
					\multicolumn{2}{|c|}{Validation}           & -     & -          & -     & 150    & 10     & 12    &1200  \\  \hline \hline
					\multirow{4}{*}[-0.6cm]{Testing}
					& Plain            & -     & -          & -     & 100   & 45     & 50   &1000    \\ \cline{2-9}
					& JPEG Compression & QF    & 50-90      & 10    & 100×5 & 45×5   & 50×5  &1000×5   \\ \cline{2-9}
					& Noise Addition   & Var   & 0.002-0.01 & 0.002 & 100×5 & 45×5   & 50×5   &1000×5  \\ \cline{2-9}
					& Resize Operation & Ratio & 0.5-0.9    & 0.1   & 100×5 & 45×5   & 50×5   &1000×5  \\ \hline \hline
					\multicolumn{2}{|c|}{All Images}           & -     & -          & -     & 3100  & 855   & 996   &27000   \\ \hline
		\end{tabular}}}
	\end{center}
	\label{tab9}
\end{table}
\subsection{Reference Methods}\label{4.3}


We only consider methods that can locate the tampered regions at the pixel level. They are color filter array (CFA) \cite{ferrara2012image}, noise inconsistency (NOI) \cite{mahdian2009using}, aligned double quantization (ADQ) \cite{lin2009fast}, error level analysis (ELA) \cite{krawetz2007picture}, RRU-Net \cite{bi2019rru}, ManTra \cite{wu2019mantra}, LSTM \cite{bappy2019hybrid}, coarse-to-refined network (C2RNet) \cite{xiao2020image}, and MAG \cite{kniaz2019point}. All of these methods can detect tampered regions at the pixel level. Some detection methods, such as those in \cite{zhou2018learning} and \cite{bayar2018constrained}, can judge only the type of forgery, so are excluded from the present comparison. The first four methods are based on traditional feature extraction. Their source codes were implemented and published by Zampoglou, \emph{et al}. \cite{zampoglou2017large}. The remaining five detection methods are based on CNNs, and their source codes are provided by the authors. The hyperparameters of the detection networks are set to the values yielding the best performances in the original papers.
\begin{table*}[tbp!]
	\setlength{\abovecaptionskip}{1 pt}
	\setlength{\belowcaptionskip}{1 pt}
	\caption{Experimental Results of Plain Splicing Forgery.}
	\begin{center}
		\scriptsize
		\renewcommand\arraystretch{1.5}
		\scalebox{1.0}{
			\setlength{\tabcolsep}{3pt}{
				\begin{tabular}{|c|c|m{1.1cm}<{\centering}|m{0.8cm}<{\centering}|m{0.8cm}<{\centering}|m{1.1cm}<{\centering}|m{0.8cm}<{\centering}|m{0.8cm}<{\centering}|m{1.1cm}<{\centering}|m{0.8cm}<{\centering}|m{0.8cm}<{\centering}|m{1.1cm}<{\centering}|m{0.8cm}<{\centering}|m{0.8cm}<{\centering}|}\hline                        & & \multicolumn{12}{c|}{ \textbf{Detection Result}}                                \\ \cline{3-14}
					&                & \multicolumn{3}{c|}{CASIA} & \multicolumn{3}{c|}{COLUMB} & \multicolumn{3}{c|}{NIST’16} & \multicolumn{3}{c|}{FR}\\ \cline{3-14}
					\multirow{-3}{*}{ \textbf{Method}} &\multirow{-3}{*}{ \textbf{Year}} & \textit{Precision}      & \textit{Recall}      & \textit{F}     & \textit{Precision}      & \textit{Recall}       & \textit{F}     & \textit{Precision}       & \textit{Recall}      & \textit{F}    & \textit{Precision}       & \textit{Recall}      & \textit{F}  \\ \hline \hline
					ELA \cite{krawetz2007picture}   &2007  & 0.086          & 0.975          & 0.158          & 0.316         & 0.961          & 0.475         & 0.141          & 0.922          & 0.315   & 0.273          & 0.709          & 0.394        \\ \hline
					NOI \cite{mahdian2009using}    &2009  & 0.149          & \textbf{0.992} & 0.258          & 0.422         & 0.997          & 0.593         & 0.202          & \textbf{0.998} & 0.336    & 0.326          & \textbf{0.993}          & 0.491      \\ \hline
					ADQ \cite{lin2009fast}    &2009   & 0.402          & 0.585          & 0.476          & 0.367         & \textbf{0.998} & 0.536         & 0.179          & 0.873          & 0.297   & 0.294          & 0.985          & 0.453       \\ \hline
					CFA \cite{ferrara2012image}   &2012  & 0.101          & 0.971          & 0.182          & 0.442         & 0.478          & 0.459         & 0.169          & 0.995          & 0.289     & 0.265          & 0.910          & 0.410     \\ \hline \hline
					RRU-Net \cite{bi2019rru}   &2019  & 0.848          & 0.834          & 0.841          & 0.918         & 0.822          & 0.867         & 0.783          & 0.782          & 0.782     & 0.859          & 0.891          & 0.875     \\ \hline
					ManTra \cite{wu2019mantra}    &2019    & 0.821          & 0.793          & 0.807          & 0.856         & 0.849          & 0.852         & 0.816          & 0.824          & 0.82      & 0.553          & 0.431          & 0.488     \\ \hline
					LSTM \cite{bappy2019hybrid}  &2019  & 0.802          & 0.783          & 0.792          & 0.831         & 0.816          & 0.823         & 0.793          & 0.785          & 0.789     & 0.462          & 0.621          & 0.533     \\ \hline
					C2RNet \cite{xiao2020image}    &2020  & 0.581          & 0.808          & 0.676          & 0.804         & 0.612          & 0.695         & 0.468          & 0.666          & 0.55     & 0.696          & 0.538          & 0.605      \\ \hline
					MAG \cite{kniaz2019point}    &2019  & -              & -              & -              & -             & -              & -             & -              & -              & -    & 0.842          & 0.910          & 0.875          \\ \hline \hline
					D-Net            &-        & \textbf{0.866} & 0.852          & \textbf{0.859} & \textbf{0.96} & 0.901          & \textbf{0.93} & \textbf{0.863} & 0.842          & \textbf{0.852} & \textbf{0.877} & 0.902          & \textbf{0.889} \\ \hline
		\end{tabular}}}
	\end{center}
	\label{tab10}
\end{table*}

\subsection{Comparative experiments and analysis}\label{4.4}

This subsection compares the performances of the proposed D-Net and several existing detection methods. Because we aim to detect splicing forgery and localize the tampered regions, we only selected the splicing forgery images in each dataset. The splicing forgery images from CASIA were randomly divided into 1350 images for training, 150 images for validation, and 100 images for testing. The ground-truth masks of each image are handcrafted. The splicing forgery images from COLUMB were resized to 757×568 and randomly divided into 125 images for training, 10 images for validation, and 45 images for testing. The splicing forgery images from NIST’16 were randomly divided into 184 images for training, 12 images for validation, and 50 images for testing. All images were resized to 512×384. Meanwhile, the splicing forgery images from FR were randomly divided into 10800 images for training, 1200 images for validation, and 1000 images for testing. All images were resized to 512×512. Moreover, for verifying and estimating the robustness of the detection methods, we created various attack cases through JPEG compression, noise attack, and resizing operation on the four testing datasets, as explained below.

\begin{itemize}
	\item[$\bullet$] JPEG compression: the splicing forgery images are saved in JPEG format with different compression quality factors (QF).
	\item[$\bullet$] Noise attack: White Gaussian noises with a mean of zero and different variances are added to the splicing forgery images.
	\item[$\bullet$] Resize operation: the splicing forgery images are scaled using a ratio factor.
\end{itemize}

As the existing traditional detection methods are based on the compression property, they only detect JPEG images. Therefore, before running the comparison experiments, all experimental images in TIFF format are converted into JPEG format with a quality factor of 100\%. The experimental images are listed in Table \ref{tab9}.

\subsubsection{Experiments of plain splicing forgery}\label{4.4.1}
\begin{figure}[htbp!]
	\begin{center}
		\includegraphics[width=0.95\linewidth]{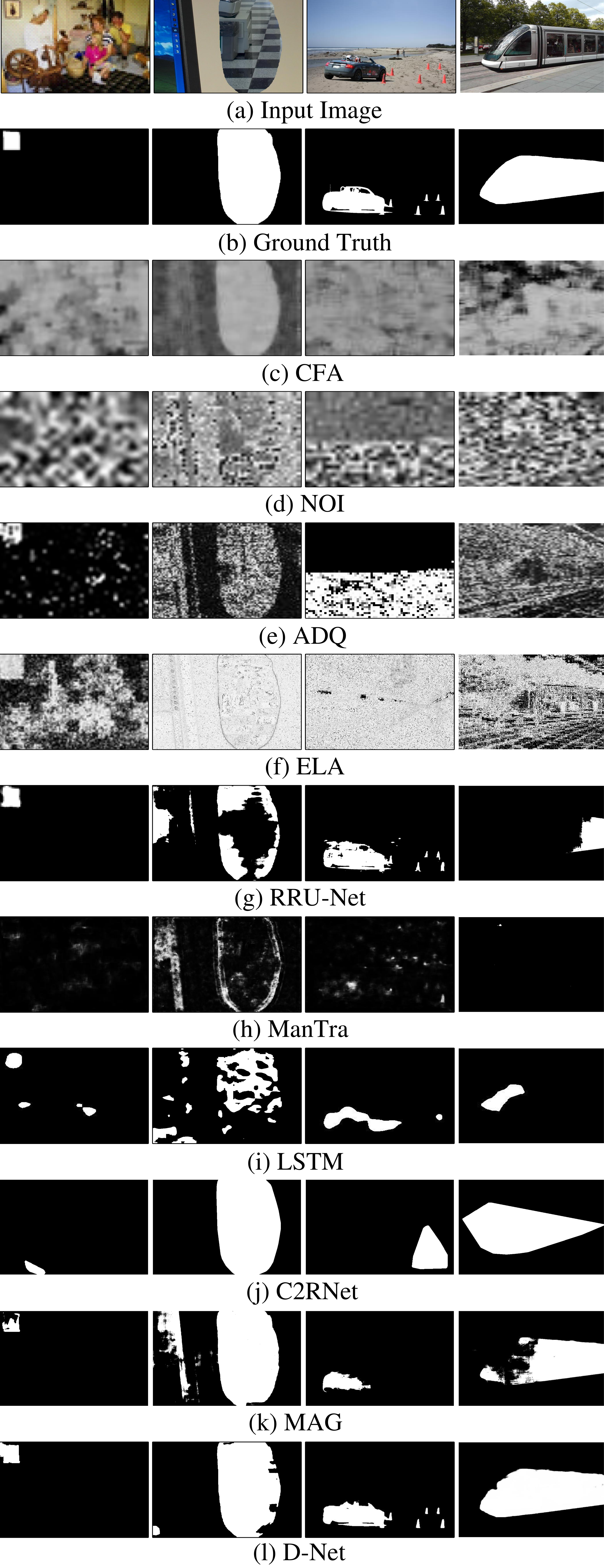}
	\end{center}
	\caption{Splicing forgery detection results of the proposed D-Net and other detection methods. The forgery images are extracted from CASIA ($1^{st}$ columns), COLUMB ($2^{nd}$ columns), NIST’16 ($3^{rd}$ columns), and FR ($4^{th}$ columns).}
	\label{fig10}
\end{figure}

This subsection compares the localization performances of the proposed D-Net and other detection methods in detecting plain splicing forgery. The localization performances of D-Net and other detection methods are assessed by the \emph{Precision}, \emph{Recall}, and \emph{F}. The performance values are listed in Table \ref{tab10}. The traditional detection methods yielded poorer \emph{Precision} and \emph{F} values than the CNN-based detection methods. The CFA, NOI, and ELA obtain very high \emph{Recall} rates because they detect almost the entire image as the tampered regions. D-Net outperforms the other CNN-based detection methods on the CASIA, COLUMB, NIST’16 datasets, and FR datasets. Moreover, we compare our method with MAG on the FR dataset only, because MAG needs class segmentations which are only provided in the FR dataset. It is especially effective on the COLUMB dataset, possibly because (unlike the other methods) D-Net extracts the directional information using a fixed encoder. Consequently, D-Net effectively locates the large, meaningless smooth tampered regions. On the FR dataset, D-Net has a relatively small improvement, which may be because the FR dataset has more samples, and other methods can learn richer traces of forgery. This further illustrates the superiority of D-Net in the case of few samples. Further subjective comparisons are shown in Fig. \ref{fig10}. Here, we show four randomly chosen examples from the four datasets. Clearly, D-Net obtains the best results among the comparison methods, even in multiple tampered regions with different scales in a forgery image.


\subsubsection{Experiments under various attacks}\label{4.4.2}
To further verify the effectiveness and robustness of D-Net, the performances of the proposed D-Net and the existing detection methods are compared under various types of attacks: noise attack, JPEG compression, and resizing operations. It is worth noting that all testing sets used for attack experiments are not included in the training set. As the LSTM method \cite{bappy2019hybrid} can only detect square images (such as 512×512- or 256×256-sized images), this method is excluded from the following experiments.

\begin{figure}[htbp!]
	\begin{center}
		\includegraphics[width=1.0\linewidth]{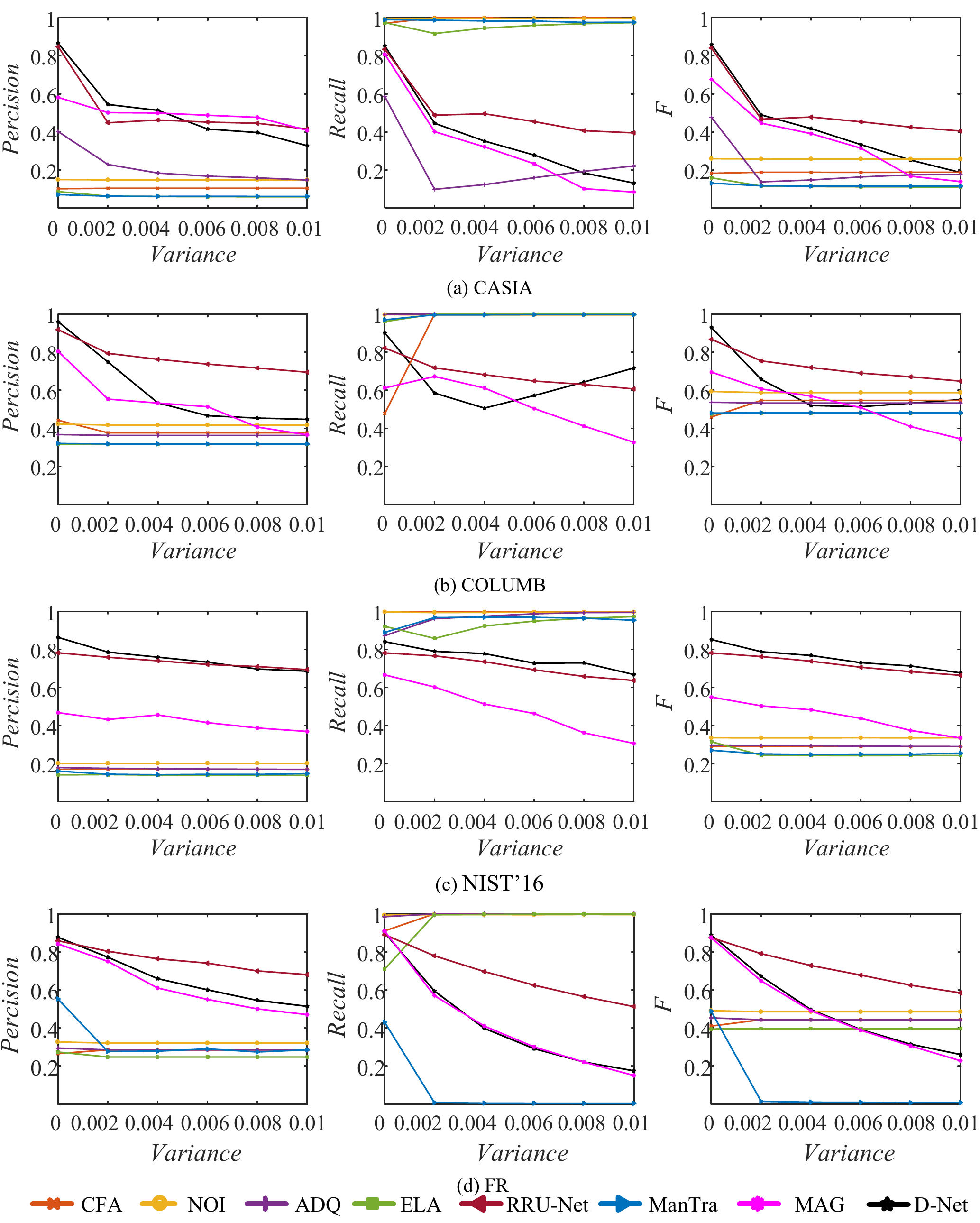}
	\end{center}
	\caption{Experiment results under the noise attacks. The first, second and third columns plot the \emph{Precision}, \emph{Recall}, and \emph{F} scores, respectively. Rows \emph{a}, \emph{b} and \emph{c} are the experimental results of the CASIA, COLUMB, NIST'16 and FR datasets, respectively.}
	\label{fig11}
\end{figure}

Adding noise to a tampered image is a general technique for hiding traces of the image manipulations. 
Therefore, an excellent forgery detection method should be sufficiently robust to noise attacks. In the first comparative experiment under an image noise attack, we estimate the robustness of D-Net and the other detection methods to noise. The experimental results are shown in Fig.
\ref{fig11}.
In Fig. \ref{fig11}, the \emph{Precision}, \emph{Recall}, and \emph{F} scores are compared under noise attacks with different variances on images from the CASIA, COLUMB, NIST'16, and FR datasets. The detection performances of the four traditional detection methods are nearly equal. Although their \emph{Precision} and \emph{F} scores are lower than in the CNN-based detection methods, their \emph{Recall} scores are elevated because traditional methods detect almost the entire image as a tampered region. Among the CNN-based detection methods, D-Net usually achieves the highest \emph{Precision} and \emph{F}, but are slightly outperformed by RRU-Net on CASIA, COLUMB and FR. D-Net achieves higher \emph{Precision} and \emph{F} on NIST’16 than the other methods. This experiment demonstrates the promising robustness of the proposed D-Net to noise attacks on all four datasets.

\begin{figure}[h!]
	\begin{center}
		\includegraphics[width=1.0\linewidth]{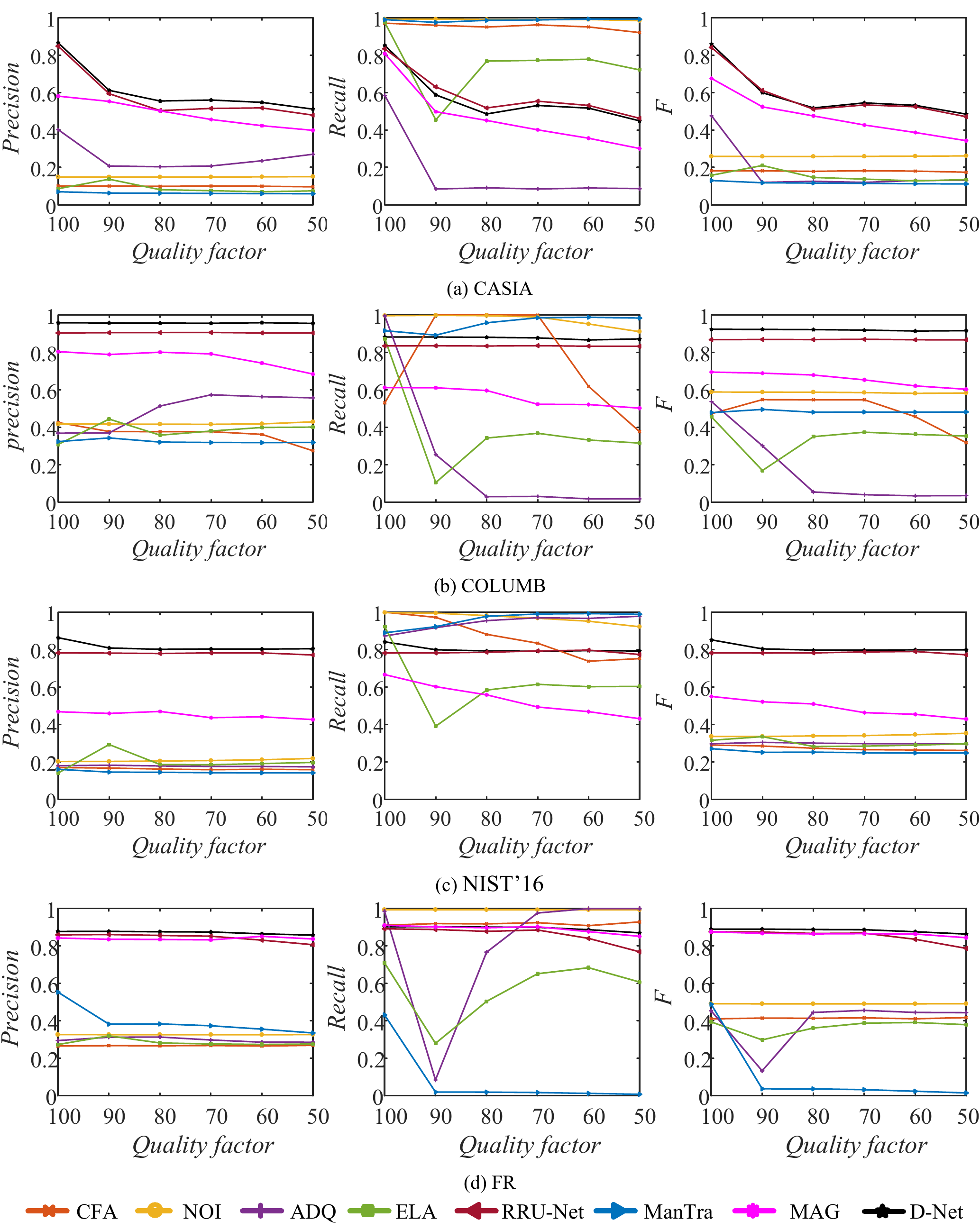}
	\end{center}
	\caption{Experimental results under JPEG compression attacks. The first, second and third columns plot the \emph{Precision}, \emph{Recall}, and \emph{F} scores, respectively. Rows \emph{a}, \emph{b} and \emph{c} are the experimental results of the CASIA, COLUMB, NIST'16 and FR datasets, respectively.}
	\label{fig12}
\end{figure}

Image compression is very common in daily life. Compression is applied to most of the images on the Internet and is also a convenient means of hiding tampering traces. Therefore, we conducted a comparative experiment under a JPEG compression attack. The experimental results are shown in Fig. \ref{fig12}.

The \emph{Precision}, \emph{Recall}, and \emph{F} scores of the different detection methods under JPEG compression attacks on images from the CASIA, COLUMB, NIST'16, and FR datasets are compared in Fig. \ref{fig12}. First, we find that the JPEG compression impacts the results of the CASIA images but exerts a very small impact on the results of the COLUMB, NIST'16 and FR images. These outcomes might be related to the compositions of the different datasets. The tampered regions of the COLUMB images are the large meaningless smooth regions, whereas the NIST’16 and FR datasets contain many samples forged from the same or very similar base images. The particularity of the tampered regions in these two datasets might resist JPEG compression attacks. Second, when the quality factor decreases from 100 to 50, the \emph{Precision}, \emph{Recall}, and \emph{F} of most of the detection methods descend dramatically, whereas the performance of D-Net remains stable. Meanwhile, the CNN-based detection methods obtain higher \emph{Precision} and \emph{F} scores than the traditional detection methods, and D-Net outperforms the other methods (similarly to the results under noise attack). In this experiment, the proposed D-Net shows promising robustness under JPEG compression attacks on all four datasets.

\begin{figure}[htbp!]
	\begin{center}
		\includegraphics[width=1.0\linewidth]{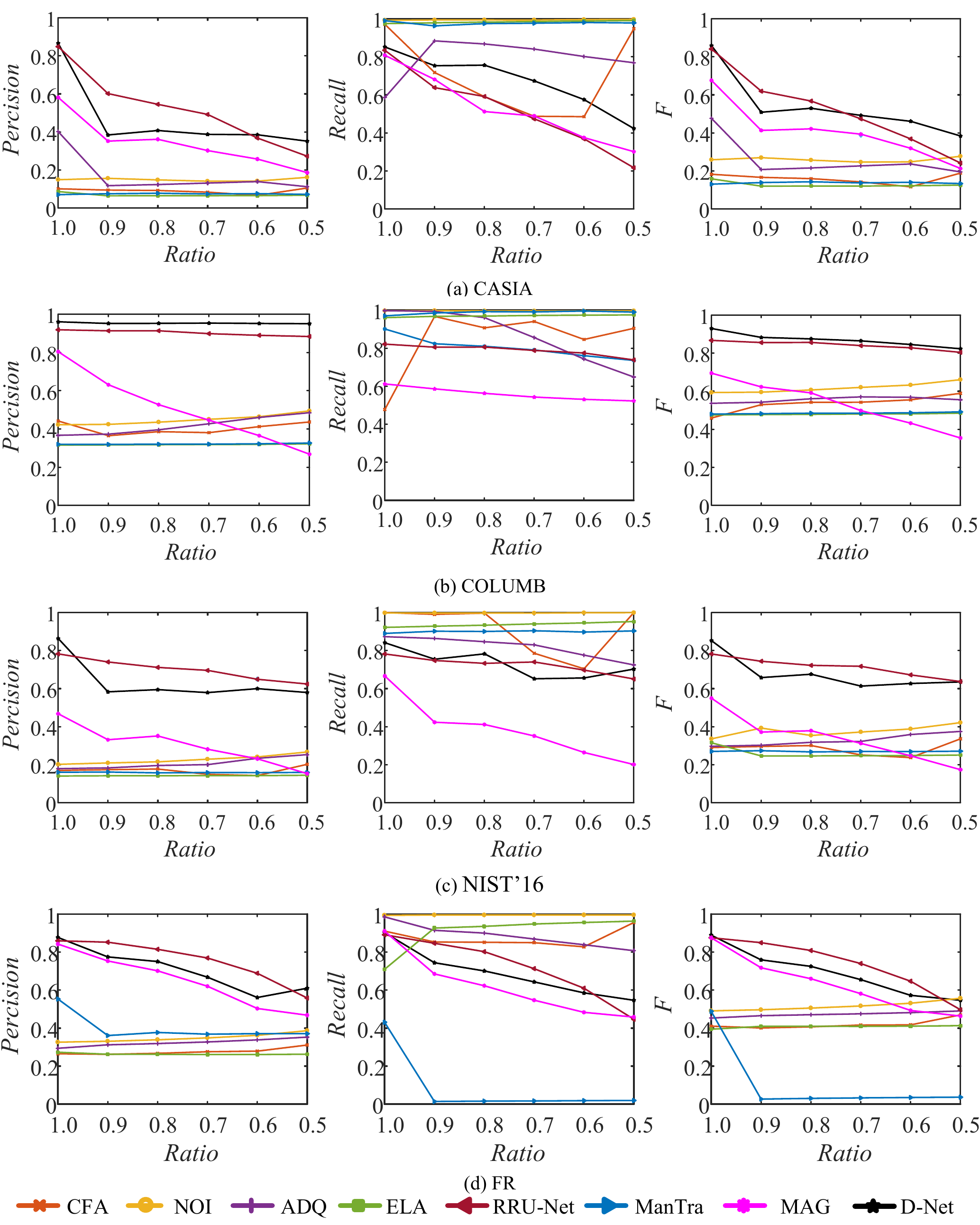}
	\end{center}
	\caption{Experimental results under the resize attacks. The first, second and third columns plot the \emph{Precision}, \emph{Recall}, and \emph{F} scores, respectively. Rows \emph{a}, \emph{b} and \emph{c} plot the experimental results of the CASIA, COLUMB, NIST'16 and FR datasets, respectively.}
	\label{fig13}
\end{figure}

The resizing operation provides a third way of hiding tampering traces. As the resizing operation usually loses some pixels, it increases the difficulty of detection. Therefore, we also compared the performances of D-Net and the existing detection methods under a resizing attack. The experimental results are shown in Fig. \ref{fig13}.

The \emph{Precision}, \emph{Recall}, and \emph{F} scores of the methods under resizing attacks with different ratios on the CASIA, COLUMB, NIST'16and FR datasets are plotted in Fig. \ref{fig13}. As indicated in the figure, D-Net achieves higher \emph{Precision} and \emph{F} scores than most of the other methods but is slightly outperformed by RRU-Net on the CASIA, NIST’16 and FR datasets. On the CASIA dataset, D-Net delivers better overall performance than the other detection methods and is only slightly worse than RRU-Net at resizing ratios above 0.8. At ratios below 0.8, D-Net consistently outperforms the other methods. On the resized COLUMB images, D-Net achieves higher \emph{Precision} and \emph{F} scores than the existing methods. This experiment confirms the robustness of the proposed D-Net to resizing attacks on all four datasets.


\section{Conclusion}\label{5}
We have proposed an end-to-end dual-encoder network (D-Net) for image splicing forgery detection, which can accurately locate the tampered regions without requiring pre- or post-processing. Owing to its structure, D-Net gains a global insight into the image splicing forgery. D-Net also maintains steady detection performance in different cases of splicing forgeries without requiring a large number of training samples or a complex pre-training process (the standard requirements of CNN-based detection methods). In the experimental comparison studies, D-Net not only outperforms the state-of-the-art methods in pixel-level detection but also presents more stable robustness under noise, JPEG compression, and resizing image attacks.


%
%

\bibliographystyle{plain}
\bibliography{refer1}

%




\end{document}